\documentclass[runningheads]{llncs}

\usepackage[T1]{fontenc}
\usepackage{graphicx}
\usepackage{xcolor}
\usepackage{comment}
\usepackage{amsmath,amssymb}
\usepackage{color}
\usepackage{url}
\usepackage{hyperref}
\usepackage{subcaption}

\usepackage{mathtools, nccmath}

\newif\ifreview
\reviewfalse

\ifreview
	\usepackage{lineno}

	\linenumbers
\fi

\begin{document}

\def\SubNumber{28}

\def\GCPRTrack{Special Track: Photogrammetry and remote sensing}

\title{PuzzleBoard: A New Camera Calibration Pattern with Position Encoding}

\ifreview
	% ANONYMOUS SUBMISSION FOR REVIEW
	% DO NOT MODIFY these for the draft version that is used for the review process.
	\titlerunning{GCPR 2024 Submission \SubNumber{}. CONFIDENTIAL REVIEW COPY.}
	\authorrunning{GCPR 2024 Submission \SubNumber{}. CONFIDENTIAL REVIEW COPY.}
	\author{GCPR 2024 - \GCPRTrack{}}
	\institute{Paper ID \SubNumber}
\else

    \author{Peer Stelldinger \and
    Nils Schönherr \and
    Justus Biermann}

    \authorrunning{P. Stelldinger et al.}

    \institute{Department of Computer Science, HAW Hamburg, Germany
    \email{peer.stelldinger@haw-hamburg.de}}
\fi

\maketitle              % typeset the header of the contribution

\begin{figure}
\centering
\includegraphics[width = 0.85\textwidth]{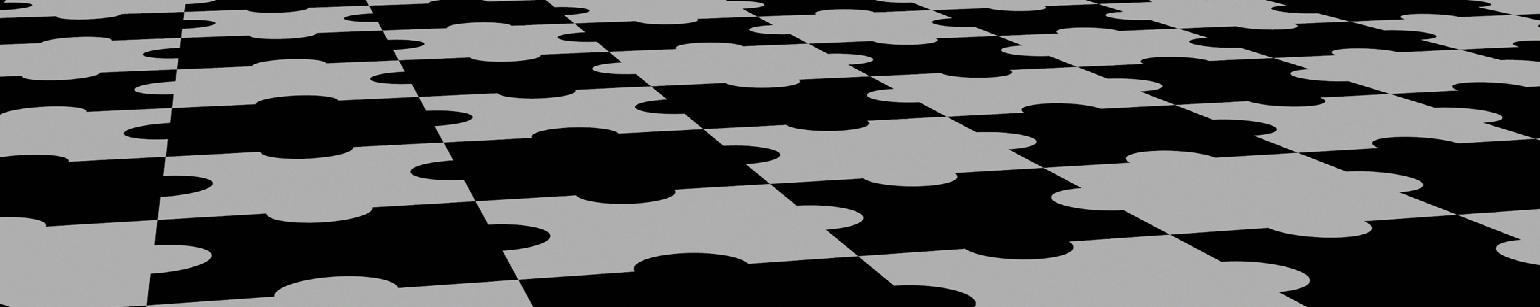}
%\caption{}
\label{fig:puzzleExample}
\end{figure}
\begin{abstract}
Accurate camera calibration is a well-known and widely used task in computer vision that has been researched for decades. However, the standard approach based on checkerboard calibration patterns has some drawbacks that limit its applicability. For example, the calibration pattern must be completely visible without any occlusions. Alternative solutions such as ChArUco boards allow partial occlusions, but require a higher camera resolution due to the fine details of the position encoding. We present a new calibration pattern that combines the advantages of checkerboard calibration patterns with a lightweight position coding that can be decoded at very low resolutions. The decoding algorithm includes error correction and is computationally efficient. The whole approach is backward compatible to both checkerboard calibration patterns and several checkerboard calibration algorithms. Furthermore, the method can be used not only for camera calibration but also for camera pose estimation and marker-based object localization tasks.

\keywords{Calibration Pattern \and \ Chessboard \and \ Checkerboard \and Fiducial Markers \and PuzzleBoard.}
\end{abstract}
\section{Introduction}

Camera calibration is a critical process in computer vision that is essential for a lot of applications in photogrammetry, 3D reconstruction, robot vision and augmented reality. After a long tradition of using 3D targets, two-dimensional checkerboard patterns have long been the go-to approach to achieving accurate calibration due to their ease of manufacture and handle. However, they have inherent limitations, especially when dealing with partial occlusions.

ChArUco boards \cite{itseez2015} combine checkerboard patterns with ArUco markers \cite{garrido2014}, allowing for partial occlusion while maintaining accuracy in camera calibration. This comes at the cost of  requiring a higher camera resolution for the position coding to be readable. This limits their usefulness in certain applications like embedded systems with low-resolution cameras. 

We present a novel calibration pattern that overcomes the limitations of existing methods. Our proposed approach combines the advantages of traditional checkerboard patterns with a lightweight position encoding scheme, allowing accurate calibration even at very low resolutions. In addition, the decoding algorithm is designed to be computationally efficient, making it suitable for embedded and/or real-time applications, and it incorporates error correction to ensure calibration even in challenging conditions like low resolution images.

We follow the idea that acquiring a high number of reference points is at least as important as detecting points with higher position accuracy, as measurement errors can be averaged out. Our approach allows this for three reasons: First, an algorithm which works at lower resolutions allows to use denser calibration patterns with more reference points. Second, the faster the algorithm, the more images of a moving calibration pattern can be used (for example with realtime feedback about the calibration accuracy). Third, being able to deal with occlusions allows to use images where only parts of the pattern are visible.

Our method ensures backward compatibility with both established checkerboard calibration patterns and several checkerboard calibration algorithms, facilitating seamless integration into existing frameworks. This means that (a) our detection algorithm also works with standard checkerboard patterns and (b) all checkerboard detection algorithms which are not affected by the code on the edges are able to decode the proposed PuzzleBoard target -- though of course both without including the position decoding. Notably, the versatility of our approach goes beyond mere calibration, serving as an alternative to ArUco-like fiducial markers when using small parts of the pattern.

One may ask the question, why such a robust calibration pattern is needed, as for many applications it is sufficient to calibrate a camera only once under laboratory conditions. The answer is, that this is not always the case. One example is cameras with variable focus and zoom lenses, as these may need to be recalibrated regularly. Another application is the joint calibration of very different cameras (with large varyity in position, orientation and/or aperture angle) in a multi camera set-up. Besides camera calibration, the advantages for high precision optical pose estimation and localization are even more obvious, as several, not necessarily flat targets can be placed in a room and each small sub-pattern allows to uniquely identify the position and orientation of a camera.

\section{Previous Work}

A very common calibration pattern design is the circle grid, which consists of a regular grid of circular blobs \cite{lanser1995,lanser1997,lenz1988,lenz1990}. The elliptical blobs can easily be detected and the position of their center point can be measured with high sub-pixel accuracy. However, the blob centroids are not invariant with respect to projective transformations and lens distortions \cite{ahn1999,heikkila2000,mallon2007,steger2017}, which needs an additional correction step \cite{steger2017}.

The still most common calibration pattern design is the checkerboard target \cite{yokobori1986,zhang2000}. Since the corners of a checkerboard appear as saddle points in an image, the detection of their position is unbiased under perspective transformations or lens distortions and can be detected with high sub-pixel accuracy without the need for bias correction -- even at extreme pose \cite{placht2014} and low resolution \cite{fuersattel2016}. 

A more recently proposed alternative is the Deltille grid \cite{ha2017}, which replaces the checkerboard corners with monkey-saddle triangular corners. Ha et al. report a 10\% increase in position accuracy per corner point \cite{ha2017} as there are three intersecting lines at each corner instead of just two. However, Deltille corners have more high-frequency components and are therefore less applicable to low-resolution images: since the incircle radius of regular triangles and squares defines the necessary sampling resolution \cite{meine2009,stelldinger2008}, and since the area of a regular triangle is $\frac34\sqrt{3}\approx 1.299$ times the area of a square with the same incircle radius, about $30\%$ more checkerboard squares than Deltille triangles can be detected in a lowest resolution image, which means approximately $2.6$ times as many corner points. Thus, the $10\%$ lower accuracy of square corner detection can be overcompensated by averaging the error over more corners ($2.6$ times more corners reduce the error to $1/\sqrt{2.6}\approx 62\%$). Deltille grids are also less applicable to extreme poses ($>65^\circ$ projection angle) than checkerboard corners \cite{ha2017}. This is even more true for higher order star shaped corners as proposed by Schöps et al.\ \cite{schoeps2020}.

All of these calibration patterns are susceptible to partial occlusion if no countermeasures are taken \cite{hillen2023}. Several alternatives have been proposed to address this shortcoming. For example, for small off-center occlusions due to clipping, it may be sufficient to add some coded marker points at the center \cite{schoeps2020,schramm2021}. However, in case of more severe occlusions, for example in case of multi-camera systems with limited overlapping field of view, or for continuous recalibration of a variable zoom and focus camera \cite{willson1994,ayaz2017}, a higher number of finder patterns in different regions of the target \cite{steger2018} or dense codes are necessary. Such higher robustness with respect to partial occlusion is provided by ArUco \cite{garrido2014} and ARTag boards \cite{fiala2008} at the cost of less accurate position information. CALTag boards \cite{atcheson2010} and ChArUco boards, as proposed in the OpenCV library \cite{itseez2015}, and similar approaches \cite{xing2017} combine the occlusion robustness of fiducial marker based calibration boards with the high position accuracy of checkerboard calibration patterns, and have thus gained traction for their robustness and versatility. However, such fiducial marker-based calibration patterns require a significantly higher image resolution than checkerboard patterns as the fiducial markers need to be readable.

Fiducial markers are not the only way to achieve position encoding. A less common approach, which to our knowledge has not yet been used for camera calibration targets, is to use 2D perfect maps, also called de Bruijn tori. A binary de Bruijn torus is a cyclic array with values 0 and 1, where each sub-pattern of a given size $(m$$\times$$n)$ occurs exactly once within one period, such that each such sub-pattern encodes a unique position. While a perfect map or de Bruijn torus contains every possible sub-pattern of the given size exactly once, a subperfect map relaxes this condition to at most once. Scheinerman is the first to use  two-dimensional generalizations of de Bruijn sequences for position encoding \cite{scheinerman2001}, but not for camera applications. Szentandrási et al.\ propose Uniform Marker Fields (UMF) \cite{szentandrasi2012}, which use binary orientable sub-perfect maps (erroneously called de Bruijn torus there) for position identification and combine this with sub-pixel accurate localization based on line fitting. Thus, their approach combines localization and identification in one pattern. The total UMF pattern consists of $92$$\times$$92$ black and white squares where each sub-pattern of size $4$$\times$$4$ is unique, even under rotation. Later, UMFs have been extended to non-binary patterns as then smaller sub-patterns identify a unique position \cite{herout2013}. However, the line fitting used in these UMF approaches requires an undistorted camera image and therefore cannot be used for camera calibration itself, but only for camera pose estimation. 

Schlüsselbauer et al. propose Dothraki \cite{schluesselbauer2021} for localising a tangible on a tabletop surface based on a binary 2D de Bruijn torus. They use a binary pattern of size $8192$$\times$$4096$ where each $5$$\times$$5$-sub-pattern is unique. Since there is no perspective distortion in their application, but only translation \cite{schluesselbauer2021} and optionally rotation \cite{schmid2022}, the decoding of the local pattern and the determination of the exact position can be done by a simple grid search with contrast maximization \cite{schluesselbauer2021} or by a simple 3-layer CNN \cite{schmid2022}. 

These codes, which are based on (sub-)perfect maps, lack a search pattern. This makes the decoding process even more challenging when distortions are not limited to rigid or perspective transformations. This explains, why common calibration targets as well as fiducial markers and 2D bar codes all use certain search patterns (for example circular blobs, ckeckerboard corners, or a black square shaped border in case of ArUco markers).

Building on this prior work, the proposed PuzzleBoard calibration pattern combines the precision and robustness of checkerboard calibration with position encoding based on a (sub-)perfect map. We combine two binary sub-perfect maps into a $4$ary map by placing their code bits at the center of the horizontal respectively vertical edges of the checkerboard squares. 

We further present a fast and robust decoding algorithm that includes error correction and supports the detection of multiple targets in one image.

By seamlessly integrating the benefits of traditional checkerboard patterns with a lightweight position coding scheme, our method overcomes the limitations of existing approaches. Furthermore, its compatibility with established algorithms ensures a smooth transition for practitioners seeking to improve the accuracy and efficiency of camera calibration procedures.

Furthermore, just as fiducial markers are used on a ChArUco board to introduce position encoding into calibration patterns, our approach could also be used for camera pose estimation and marker-based object localization tasks. For example, small sub-patterns of the PuzzleBoard pattern provide easy to read  fiducial markers. One could even think of covering an entire floor with a PuzzleBoard pattern as a localization means for autonomous robots or drones. 

The remaining paper is organized as follows:
In Section~\ref{sec1}, we introduce the encoding scheme and use this to define the PuzzleBoard calibration pattern design. In Section~\ref{sec2} we present the decoding algorithm in detail before we evaluate its performance in Section~\ref{sec3}, followed by the conclusion in Section~\ref{sec4}.

\section{Position Encoding} \label{sec1}

A sub-perfect map of type $(M,N;\; m,n)_k$ is a cyclic two-dimensional array of shape $(M,N)$ of letters from an alphabet of size $k$, where every possible $(m,n)$-pattern of the letters occurs at most once \cite{mitchell1994}. If every possible $(m,n)$-pattern occurs exactly once, it is called a perfect map\cite{paterson1994}, or a de Bruijn torus \cite{cook1988,hurlbert1993}. De Bruijn tori are thus a natural two-dimensional generalization of de Bruijn sequences \cite{debruijn1946}.

\begin{figure}
\centering
\begin{subfigure}{\textwidth}
\centering
\includegraphics[width = \textwidth]{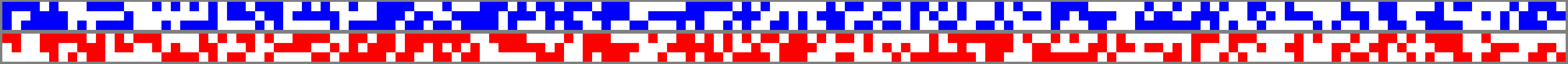}
\caption{Top (blue): Sub-perfect map $\mathrm{A}$. Bottom (red): Rotated sub-perfect map $\raisebox{\depth}{\rotatebox{90}{B}}$. In each map, each $3$$\times$$3$-pattern (with wrap around) is unique.}
\end{subfigure}
\begin{subfigure}{0.325\textwidth}
\centering
\includegraphics[width = \textwidth]{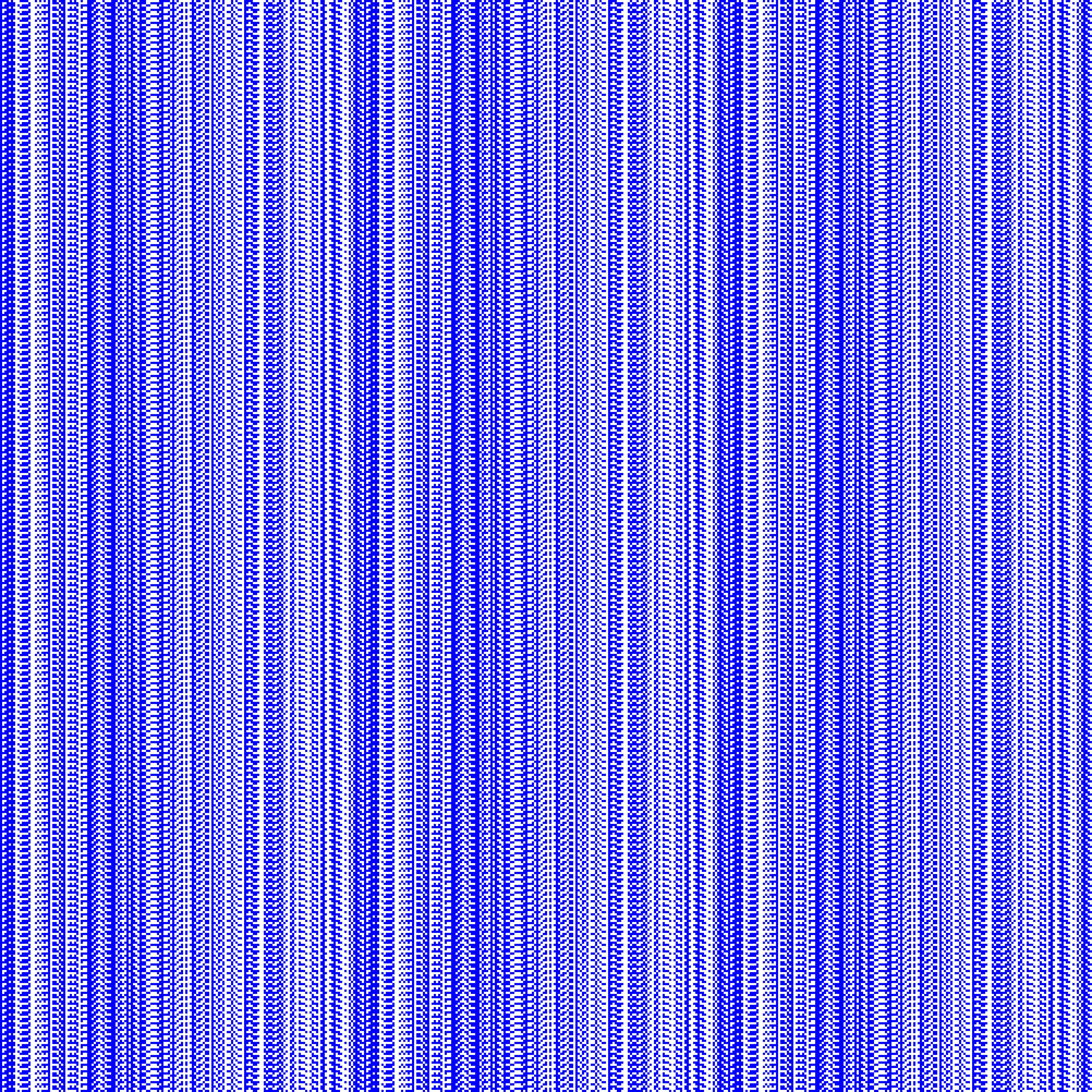}
\caption{$167$$\times$$3$ repetitions of $\mathrm{A}$.}
\end{subfigure}
\begin{subfigure}{0.325\textwidth}
\centering
\includegraphics[width = \textwidth]{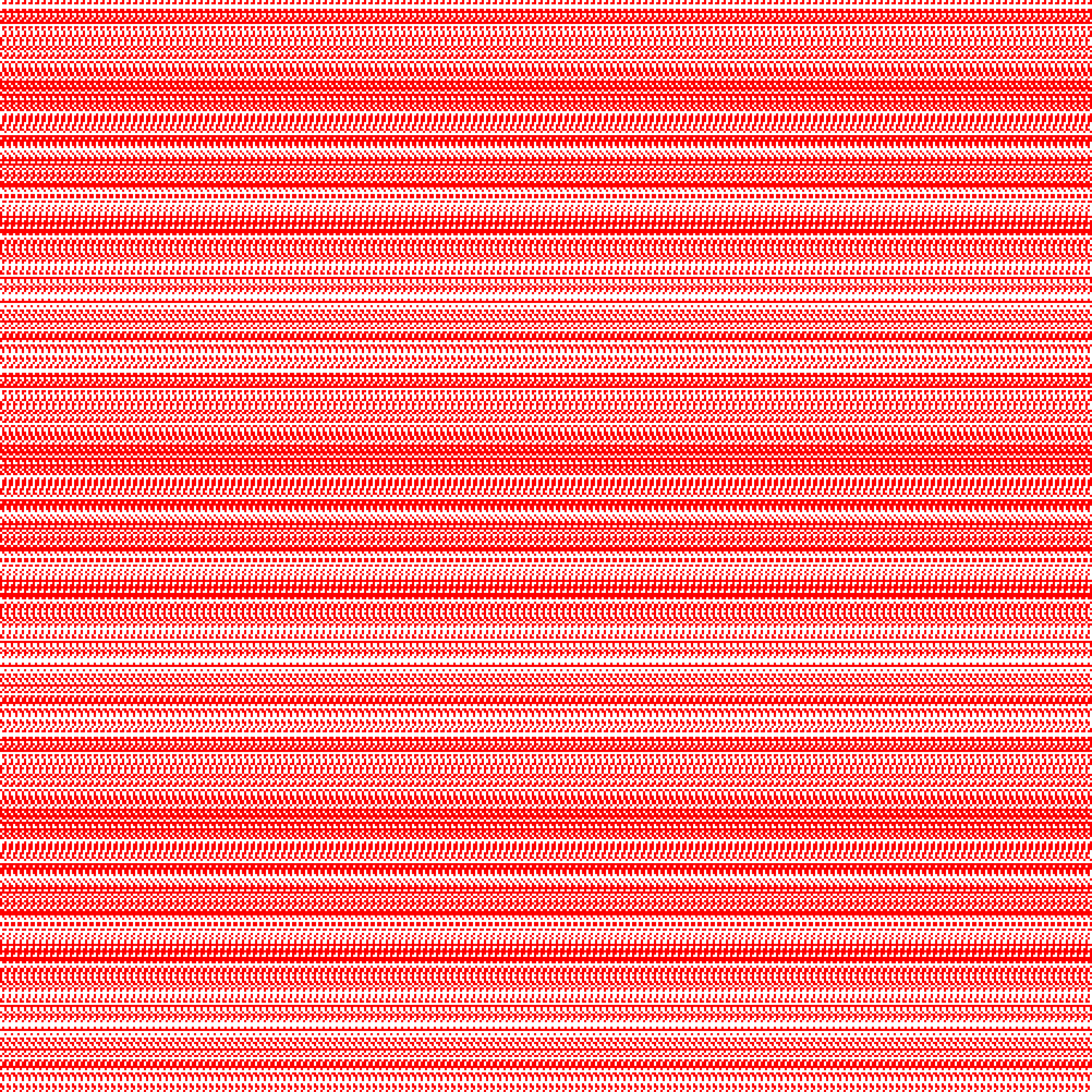}
\caption{$3$$\times$$167$ repetitions of $\mathrm{B}$.}
\end{subfigure}
\begin{subfigure}{0.325\textwidth}
\centering
\includegraphics[width = \textwidth]{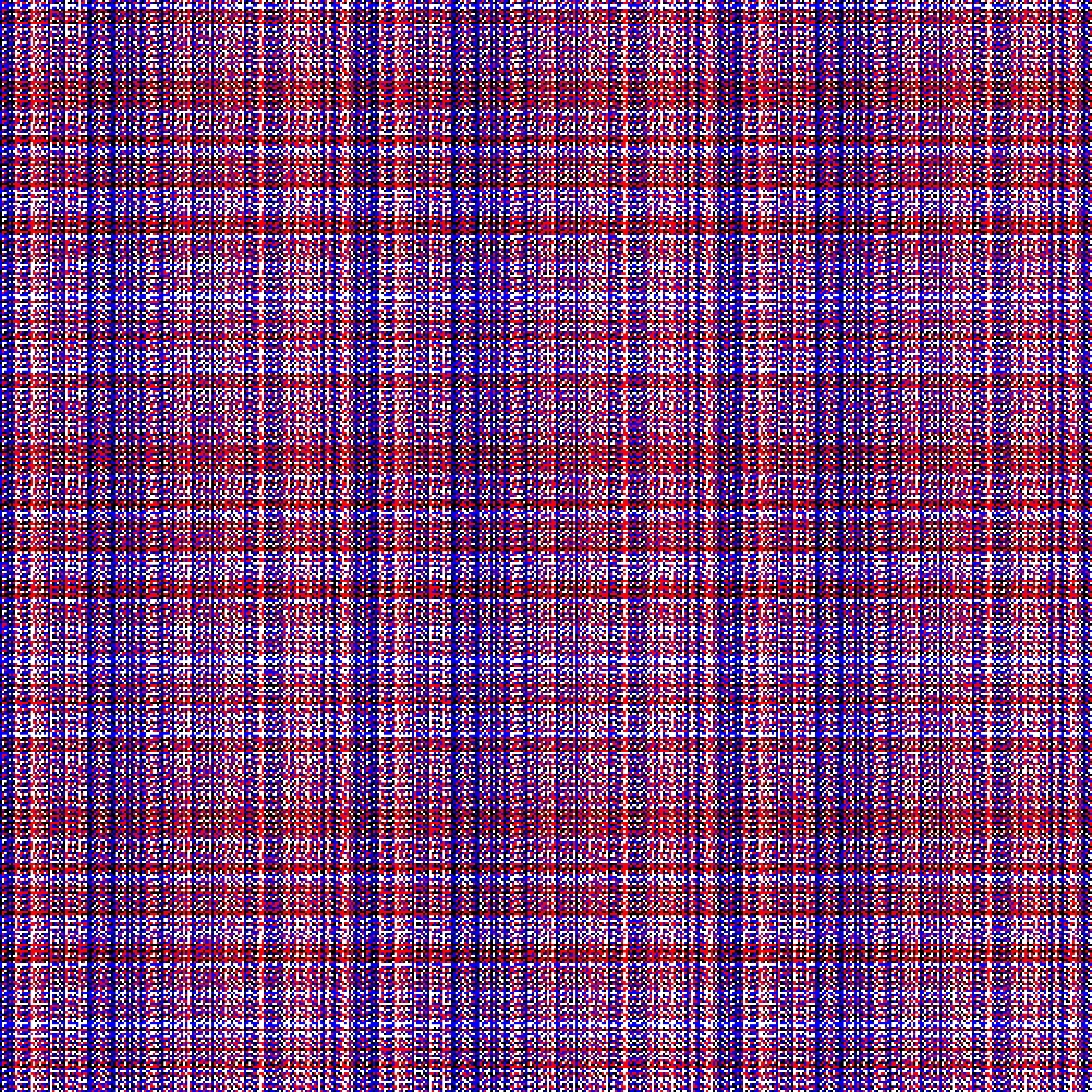}
\caption{Combined 4ary code.}
\end{subfigure}
\caption{Construction of a sub-perfect map of type $(501,501;\; 3,3)_4$ based on two $(3,3)_2$ de Bruijn rings. Every $3$$\times$$3$ pixel pattern in (d) is unique.}
\label{fig:basecodes}
\end{figure}

For position encoding tasks, the map does not necessarily have to be perfect: any sub-perfect map will suffice as long as it is almost perfect and an efficient decoding method is given \cite{stelldinger2024}. We generate a sub-perfect map of type $(501,501;\; 3,3)_4$ by superposing the cyclic repetition of a sub-perfect map $\mathrm{A}$ of size $(3,167;\; 3,3)_2$ with the cyclic repetition of a second sub-perfect map $\mathrm{B}$ of size $(167,3;\; 3,3)_2$, see Figure~\ref{fig:basecodes}. This construction process follows the method given in \cite{stelldinger2024}, which is based on so-called $(3,3)_2$ de Bruijn rings. 

\begin{figure}
\centering
\begin{subfigure}{0.19\textwidth}
\centering
\includegraphics[width = \textwidth]{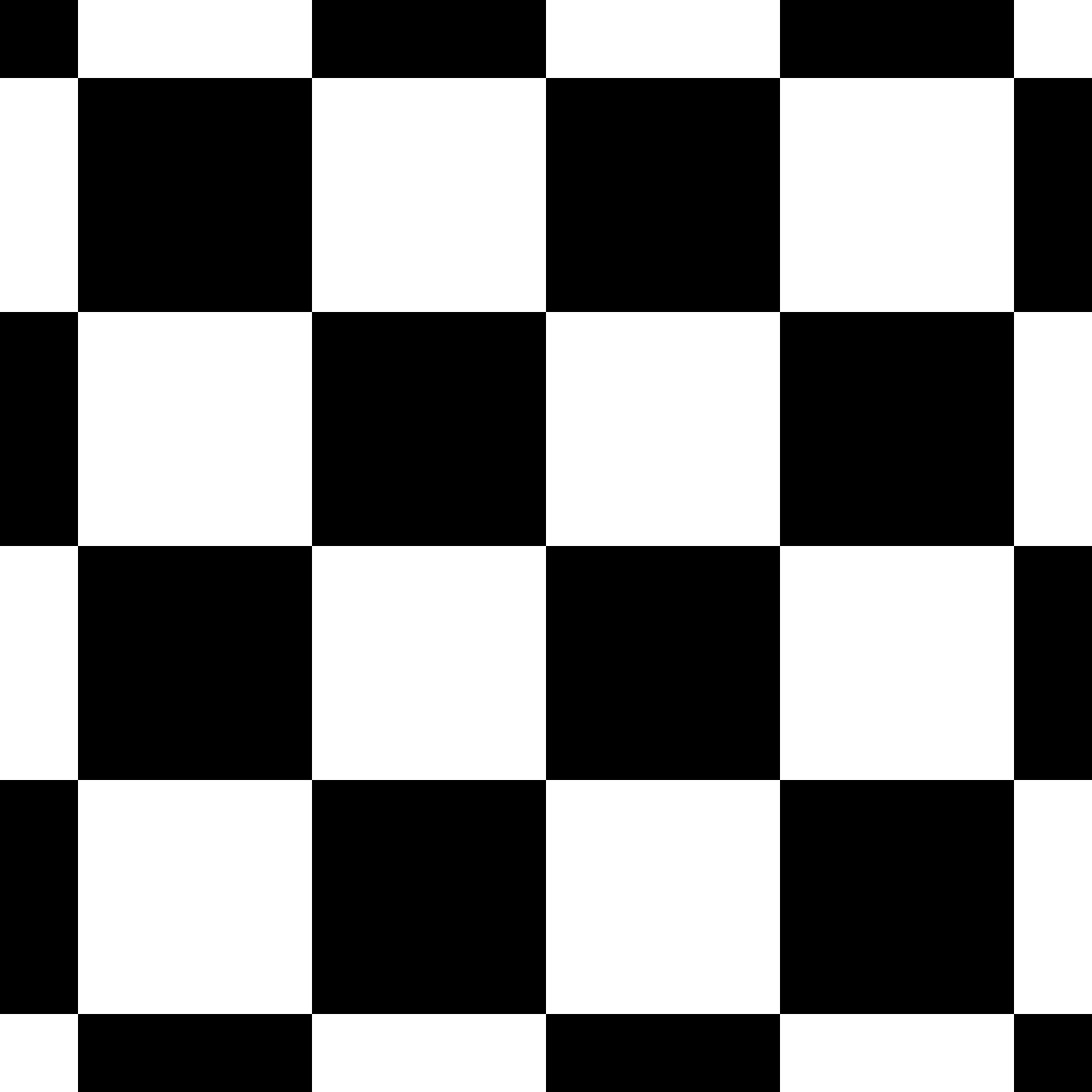}
\caption{checkerb.}
\end{subfigure}
\begin{subfigure}{0.19\textwidth}
\centering
\includegraphics[width = \textwidth]{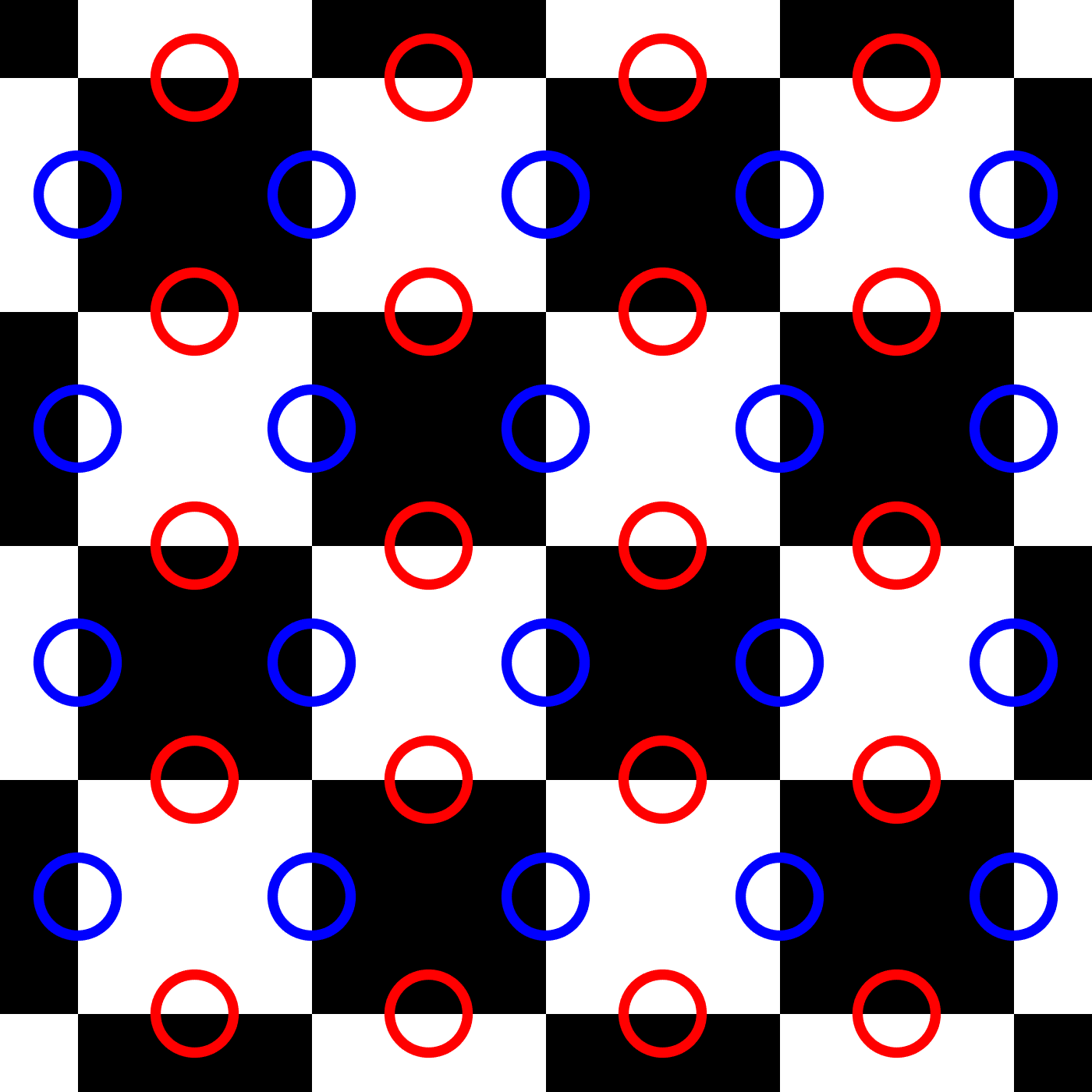}
\caption{bit positions}
\end{subfigure}
\begin{subfigure}{0.19\textwidth}
\centering
\includegraphics[width = \textwidth]{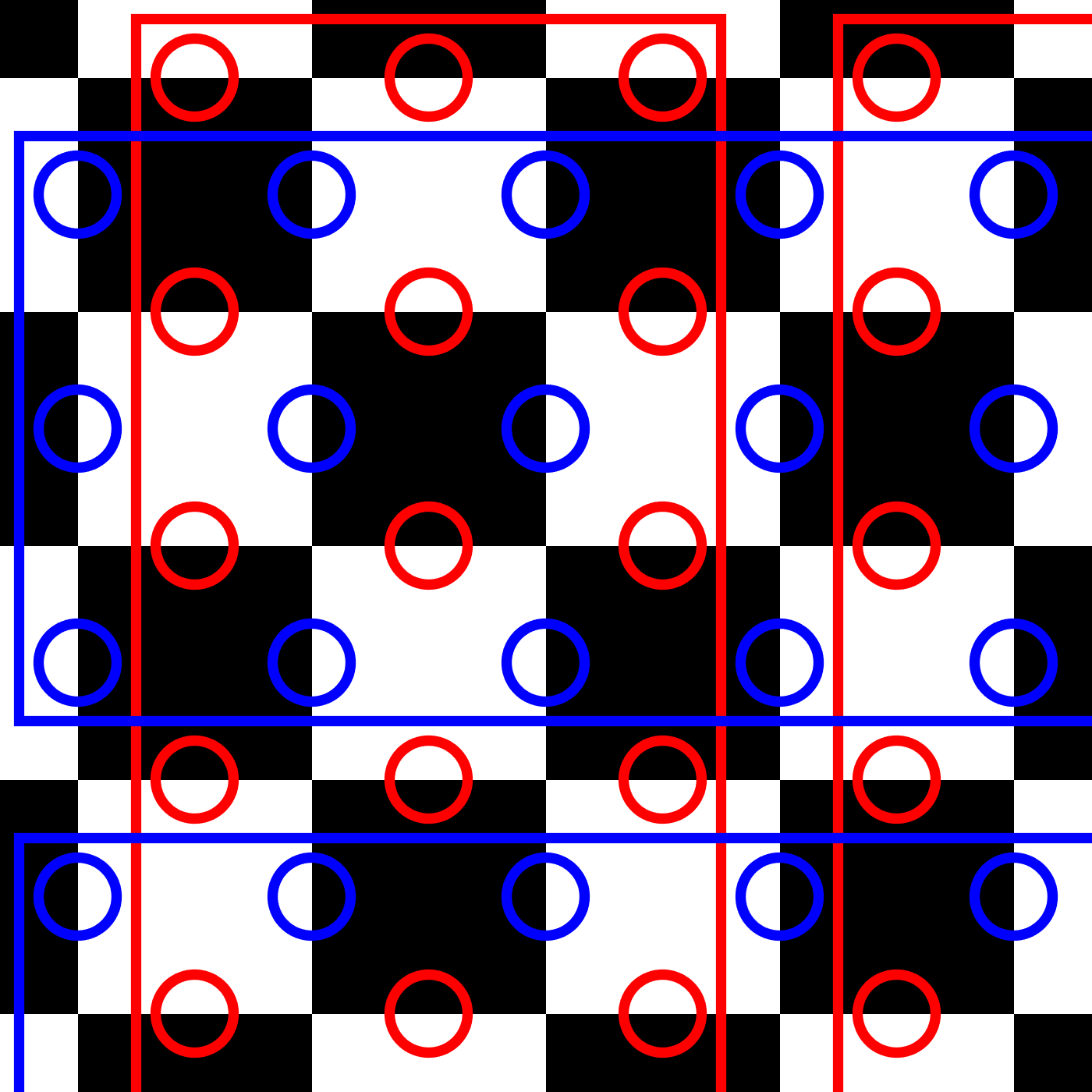}
\caption{codes}
\end{subfigure}
\begin{subfigure}{0.19\textwidth}
\centering
\includegraphics[width = \textwidth]{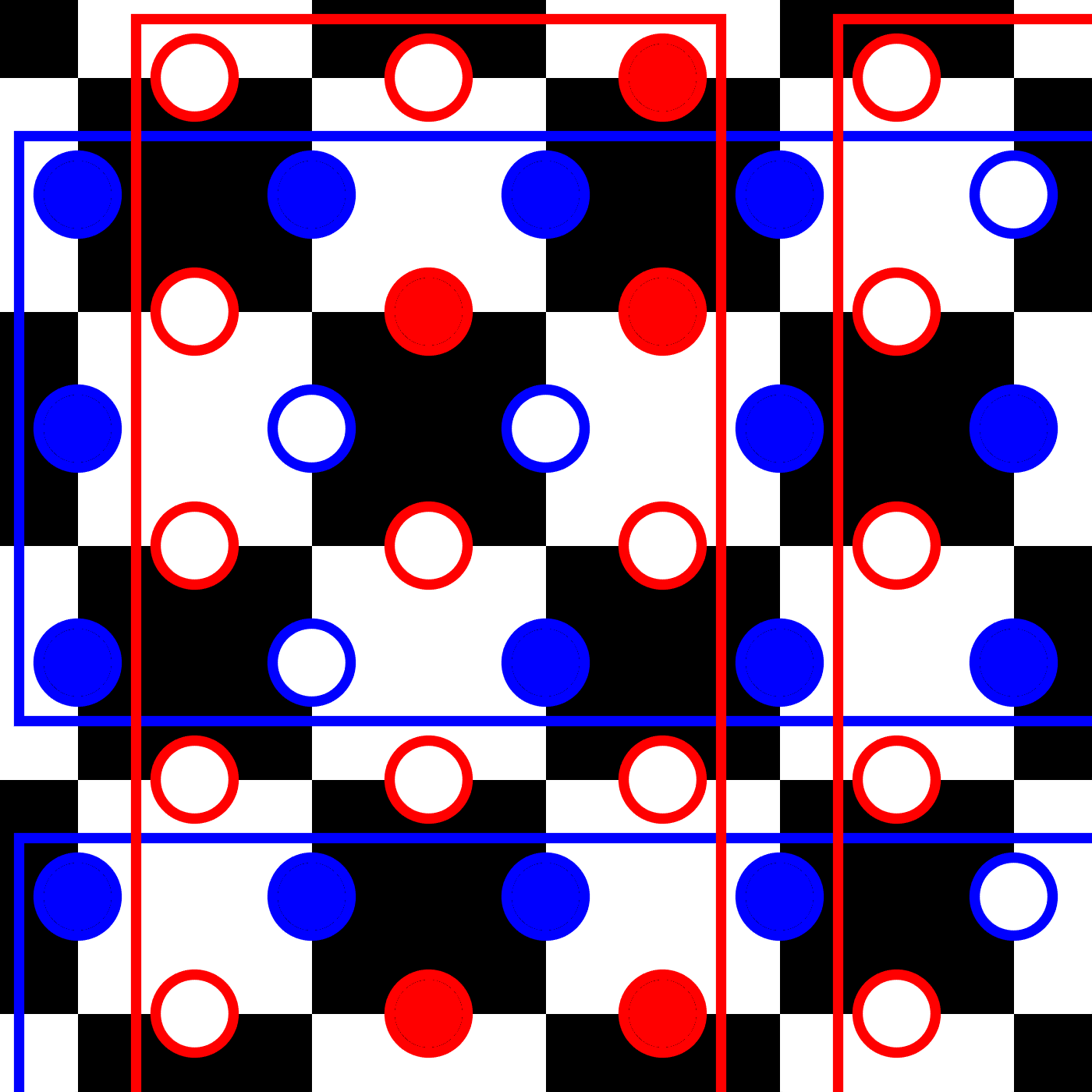}
\caption{bit values}
\end{subfigure}
\begin{subfigure}{0.19\textwidth}
\centering
\includegraphics[width = \textwidth]{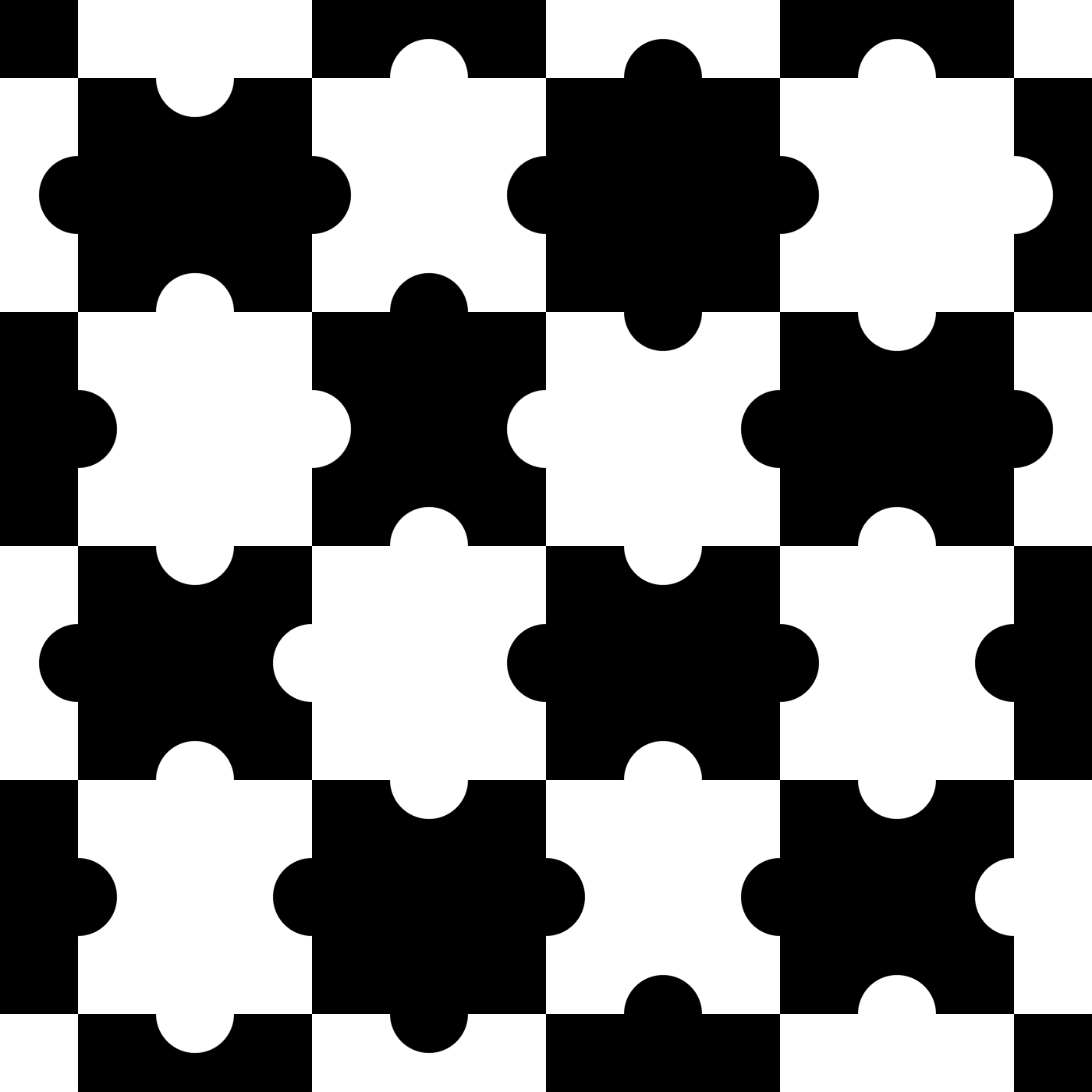}
\caption{PuzzleBoard}
\end{subfigure}
\caption{Starting with a checkerboard (a), circles with diameter $\frac{1}{3}$ of the edge length are placed on the {\color{red} horizontal} and {\color{blue} vertical} edges (b), grouped in blocks of size {\color{red} $167$$\times$$3$} and {\color{blue} $3$$\times$$167$} (c), and the bit patterns $\mathrm{A}$ and $\mathrm{B}$ are added (d) to get the PuzzleBoard (e). }
\label{fig:pattern}
\end{figure}

\begin{figure}
\centering
\begin{subfigure}{0.19\textwidth}
\centering
\includegraphics[width = \textwidth]{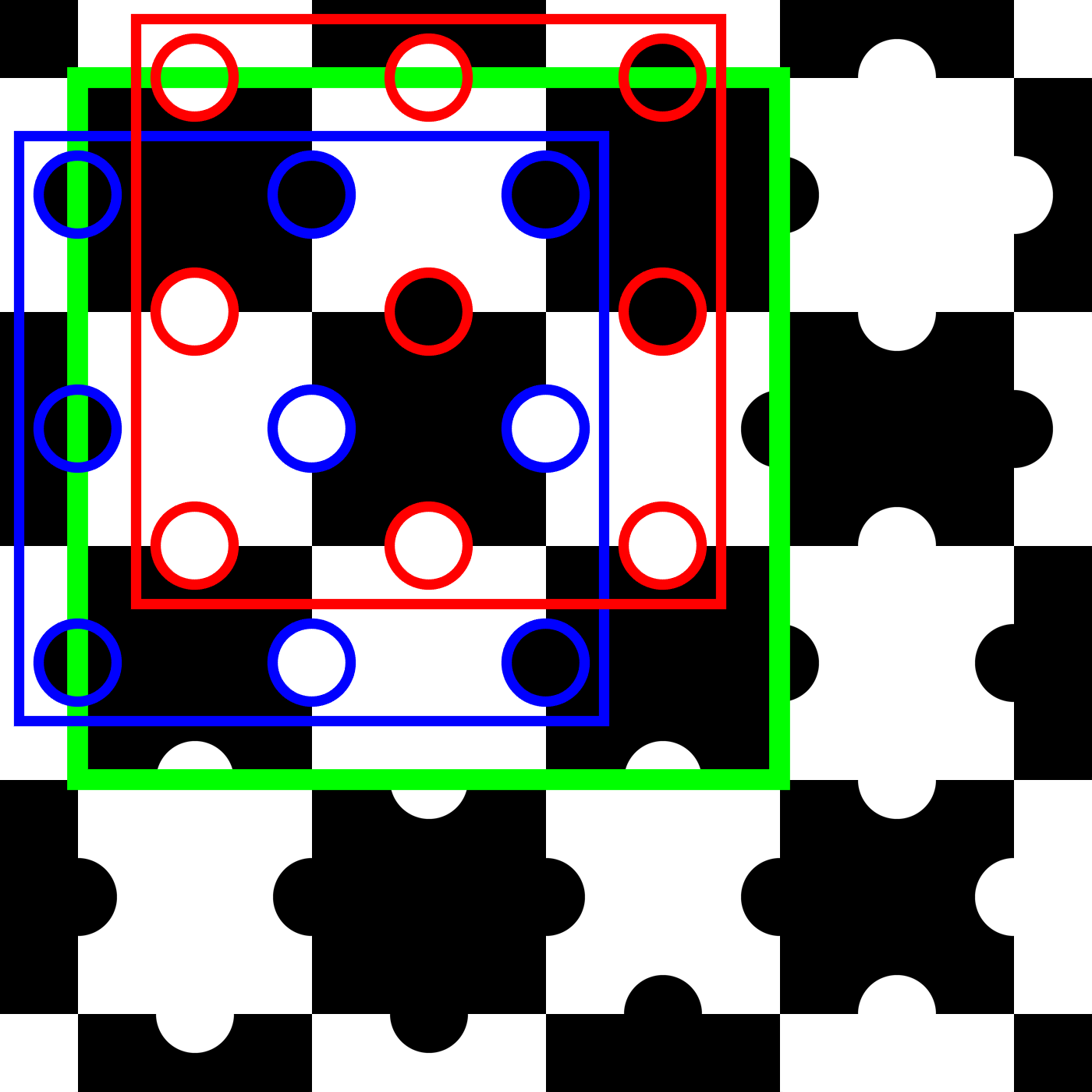}
\caption{18 bits}
\end{subfigure}
\begin{subfigure}{0.19\textwidth}
\centering
\includegraphics[width = \textwidth]{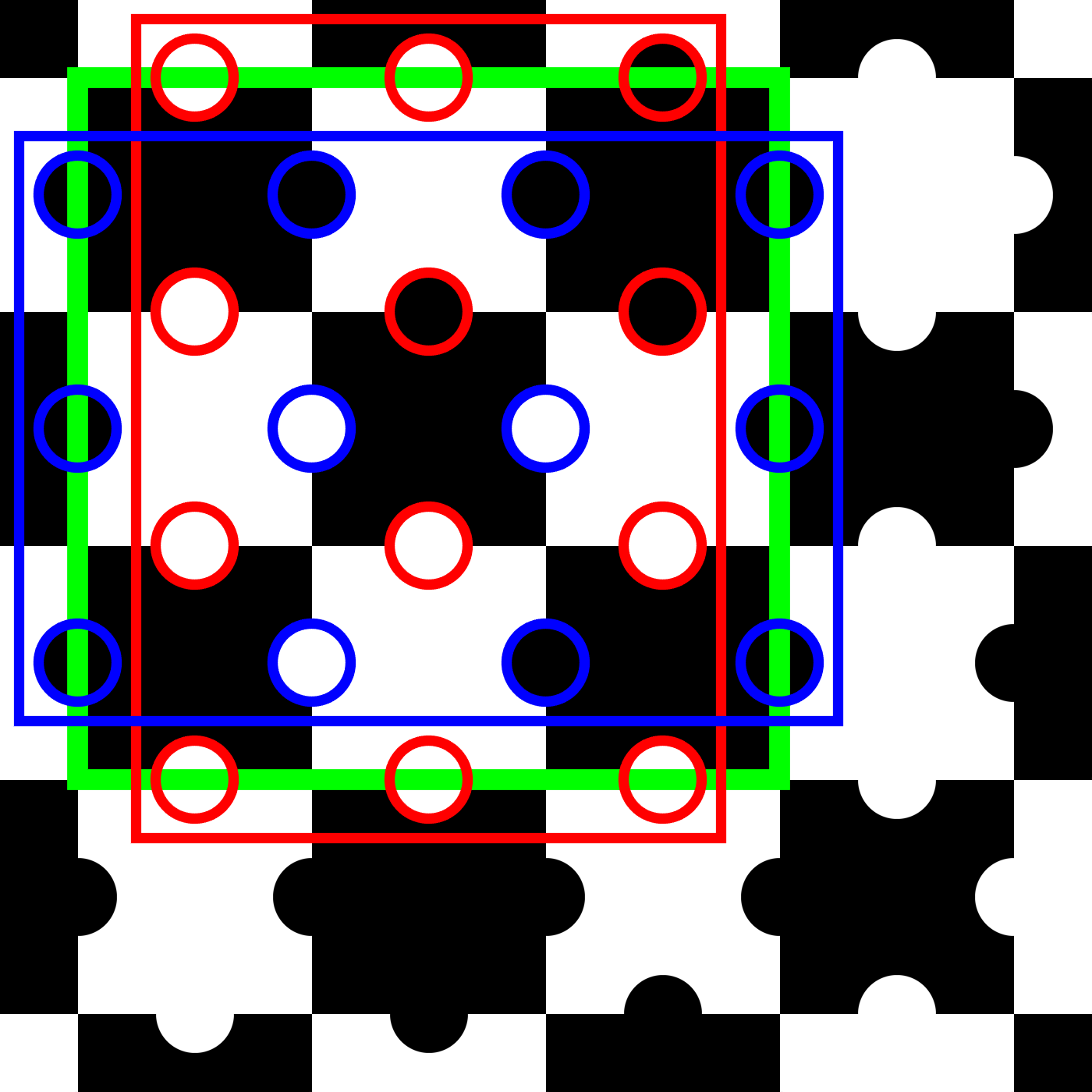}
\caption{24 bits}
\end{subfigure}
\begin{subfigure}{0.19\textwidth}
\centering
\includegraphics[width = \textwidth]{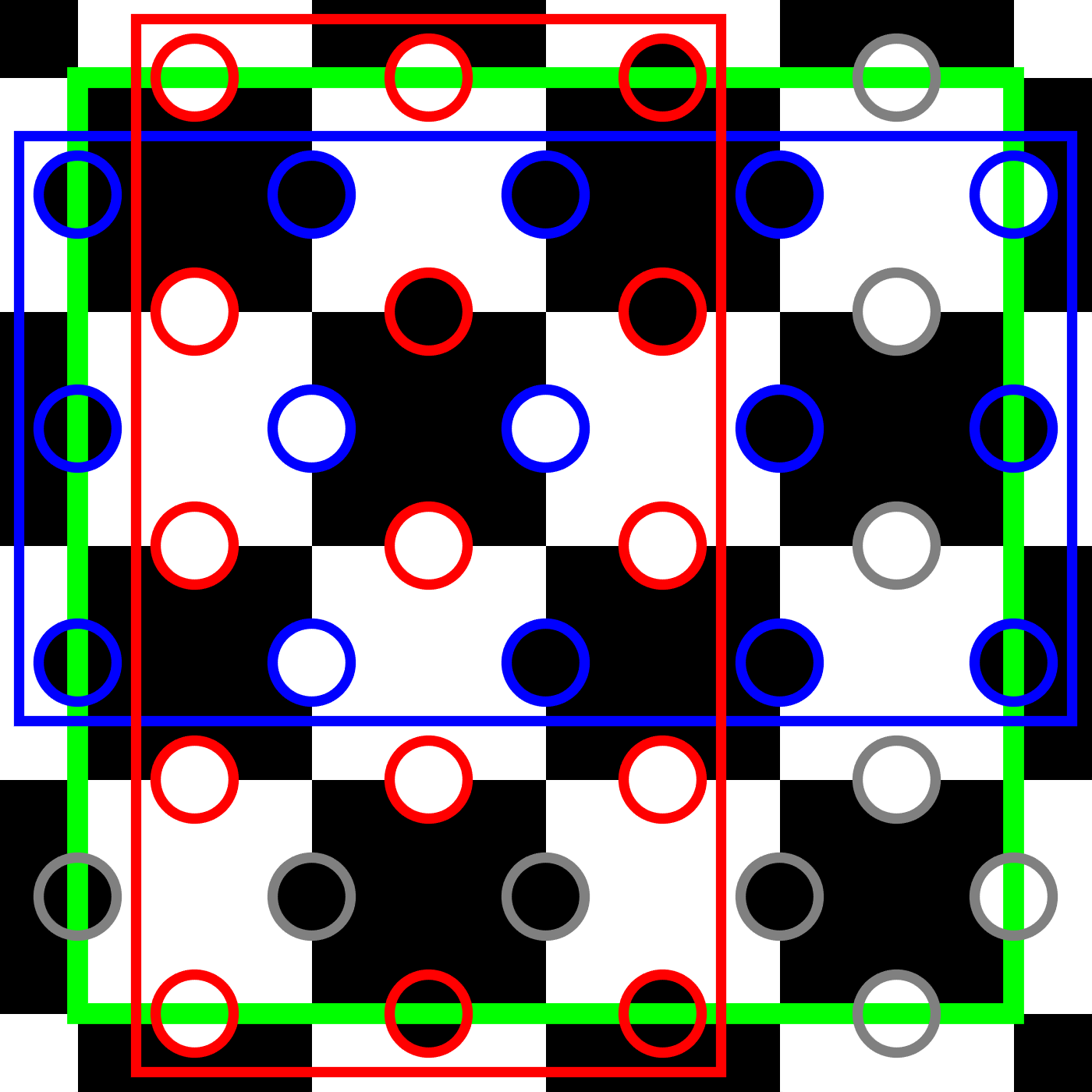}
\caption{30 bits}
\end{subfigure}
\caption{The 18 bits of each $3$$\times$$3$ pattern (a) are unique if the orientation is known. Using all 24 bits of a $3$$\times$$3$ pattern (b) adds rotational uniqueness for $99,33\%$ of the patterns, while $4$$\times$$4$ patterns (c) are always unique under rotation.}
\label{fig:bits}
\end{figure}

In contrast to \cite{stelldinger2024}, we do not combine the binary alphabets of the two patterns $\mathrm{A}$ and $\mathrm{B}$ to a 4-character alphabet, but use the patterns separately to add one bit of information at the center of each horizontal or vertical edge of a checkerboard of $501$$\times$$501$ pieces. Figure~\ref{fig:pattern} shows the construction. As a result, each checkerboard piece carries an average of 2 bits (actually it is 4 bits, but each bit of information is shared by two neighboring pieces). Thus, by assigning e.g.\ the top and the left bit to a PuzzleBoard piece, each sub-pattern of $3$$\times$$3$  pieces carries a unique 18 bit code (see Figure~\ref{fig:bits}a).

\begin{figure}
\centering
\includegraphics[width = 0.65\textwidth]{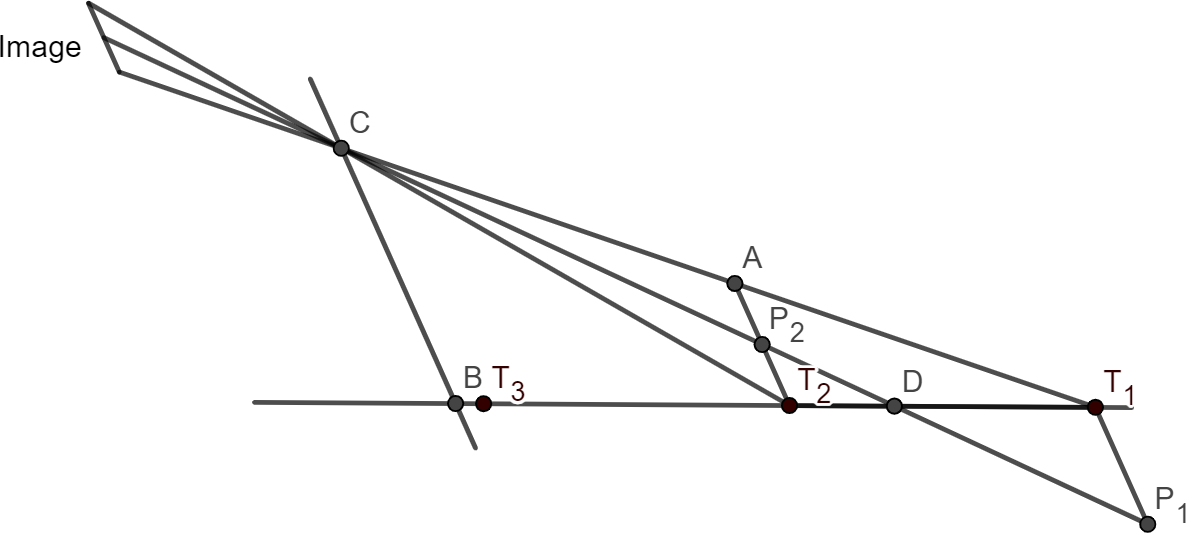}
\caption{A PuzzleBoard pattern with neighboring PuzzleBoard corners $T_1,T_2$ is projected onto an image by a pinhole camera with camera center $C$. The lines $CB$, $T_2A$ and $T_1P_1$ are parallel to the image plane. The point $D$, whose projection onto the image plane lies at the center between the projection of two neighboring PuzzleBoard corners $T_1$ and $T_2$ divides $\overline{T_1T_2}$ into two segments of length ratio $|T_2D|/|DT_1|=|T_2P_2|/|T_1P_1|=|AP_2|/|T_1P_1|=|CA|/|CT_1|$. If $T_2$ is further away from the camera plane $CB$ than from $T_1$, then $|BT_2|>|T_3T_2|$ for a point $T_3$ with $|T_3T_2|=|T_2T_1|=\frac{1}{2}|T_3T_1|$. Then it follows, that $|CA|/|CT_1|=1-|AT_1|/|CT_1|=1-|T_2T_1|/|BT_1|\geq 1-|T_2T_1|/|T_3T_1|=\frac{1}{2}$. Thus, under the weak assumption that the next corner point $T_3$ is still in front of the camera, the center point of the image of a PuzzleBoard edge is the projection  of a point $D$ which lies inside the second third of the edge $T_1T_2$.}
\label{fig:camera}
\end{figure}

Representing the code as circles on the centers of the checkerboard edges has a simple benefit: reading the code means just to check the gray value at the centers between neighboring checkerboard corner points. By setting the diameter of the circle to one third of the edge length, so that it covers the middle third of the edge, it is guaranteed, that the center between two neighboring corner points lies inside this circle, even under extreme perspective distortion (see Figure~\ref{fig:camera}).

Note, that the sub-perfect map guarantees the uniqueness of each $3$$\times$$3$-sub-pattern only if its orientation is known, which cannot be assumed in general. Unfortunately, there is no known method for effectively generating so-called 4-orientable de Bruijn tori \cite{burns1992,szentandrasi2012}, i.e. sub-perfect maps, whose local sub-patterns are unique with respect to both translation and rotation. Szentandrási et al.\ \cite{szentandrasi2012} use a parallelized genetic algorithm on a 1000 node supercomputer to construct non-cyclic orientable sub-perfect maps including one of size $11$$\times$$11$ for $3$$\times$$3$ sub-patterns and one of size $92$$\times$$92$ for $4$$\times$$4$ sub-patterns. We use a different approach: the construction method given in \cite{stelldinger2024} allows to generate any possible $(3,3)_2$ de Bruijn ring, and any two such rings can be combined to obtain a sub-perfect map of type $(501,501;\; 3,3)_4$. While these maps are not 4-orientable with respect to their $3$$\times$$3$ sub-patterns, we used a stochastic hill-climbing search to find two maps whose combination minimises the number of orientation collisions when looking at a slightly larger neighborhood than $3$$\times$$3$. With the proposed map, using all 24 edges of $3$$\times$$3$ PuzzleBoard pieces (see Figure~\ref{fig:bits}b), $99,33\%$ of all $3$$\times$$3$ such local patterns are unique under orientation. Using all 40 edges of a larger pattern of $4$$\times$$4$ PuzzleBoard pieces (see Figure~\ref{fig:bits}c), all these patterns are unique under orientation. Note, that in the latter case, there is only 30 bits of information, as the remaining bits are repetitions.

The highly repetitive structure of the bit patterns allows for effective error correction: As each bit is repeated every three rows or columns, its correct value can be derived by majority voting when more rows and columns are visible. The larger the PuzzleBoard, the more robust its position coding becomes. 

The large size of the PuzzleBoard pattern allows to simultaneously use different parts of it for different use cases: Some parts may be used for calibration boards of different size (see Figure~\ref{fig:threetargets}), some may be used as markers for the identification and tracking of mobile objects, and others may cover the floor for high precision camera localization and pose estimation. In a closed environment, the specific task could be automatically derived from the detected sub-pattern. In the rare case, that the $501$$\times$$501$ PuzzleBoard pattern is not large enough for a certain task, the approach can easily be extended to larger base patterns. For example, when using  $(4,4)_2$ de Bruijn rings for constructing a PuzzleBoard, the total pattern would be $65.276$$\times$$65.276$ puzzle pieces large with every sub-pattern of size $4$$\times$$4$ being unique \cite{stelldinger2024}.

%\[
%\left[ \begin{array}{c|cccc}
%& \mathrm{A} & \raisebox{\depth}{\rotatebox{180}{A}} & \mathrm{B} & \raisebox{\depth}{\rotatebox{180}{B}} \\
%\mathrm{A} & 53(501) & 93 & 21 & 19 \\
%%\raisebox{\depth}{\rotatebox{180}{A}} & 93 & 53(501) & 19 & 21 \\
%\mathrm{B} & 21 & 19 & 58(501) & 105 \\
%%\raisebox{\depth}{\rotatebox{180}{B}} & 19 & 21 & 105 & 58(501) \\
%\end{array}\right]
%\]

%Thus, the maximal number of equal bits at some wrong position is 105 bits when code B is cross correlated to its upside down version and 93 bits when code A is cross correlated to its upside down version, while there are 501 equal bits each for the correct position. In this extreme case, up to 197 arbitrary bit errors for code B and up to 203 arbitrary bit errors for code A can be corrected, as $(501-105)/2=198$ and $(501-93)/2=204$. This means error rates of $39.3\%$ and $40.5\%$. Note, that this means, that approximately $40\%$ of the bits are allowed to be decoded incorrectly \emph{after} averaging over all repetitions and thus the error rate per edge can be even higher.   

\section{Decoding Algorithm} \label{sec2}

The decoding algorithm consists of the four steps (1) checkerboard corner detection, (2) neighbor identification, (3) grid reconstruction and (4) position decoding. All four steps are optimised for both computational time and robustness to coarse resolutions.

\subsection{Checkerboard Corner Detection}
Given a sufficiently high image resolution and low lens distortion, the most accurate way to detect checkerboard corners is to fit lines to the edges and calculate the intersection. However, with low resolution or significant distortion, reliable straight line fitting is not possible and local detection algorithms must be used. A common choice is to search for negative maxima of the determinant $\det(H)$ of the Hessian $H={\left( \begin{array}{cc}f_{xx}&f_{xy}\\f_{xy}&f_{yy}\end{array}\right)}\textrm{,}$ since a checkerboard corner appears as a saddle point in the image \cite{chen2005}. However, as such a detector gives many false positives, we reduce the number of false detections by weighted subtraction of the squared Laplace $\mathrm{trace}(H)^2=(f_{xx}+f_{yy})^2$, which is zero at saddle points. So we look for local positive maxima of
\begin{eqnarray*}
    s &=&-\det(H)-k\cdot\mathrm{trace}(H)^2\\
    &=&f_{xy}^2-f_{xx}f_{yy}-k(f_{xx}+f_{yy})^2 \mathrm{.}
\end{eqnarray*}
By default, we use $k=1$. Note, that this filter looks similar but is fundamentally different from the commonly used Harris corner detector $\det(M)-k\cdot\mathrm{trace}(M)^2$ \cite{harris1988}, which uses the structure tensor $M$ instead of the Hessian $H$ and is unable to distinguish between checkerboard corners and circular blobs, as both appear as local maxima of the autocorrelation function. An additional test on the centrosymmetric property \cite{liu2016} removes further false detections.

At the local maxima of $s$, we perform a sub-pixel corner position refinement by computing the grayscale centroid \cite{shortis1994} of non-negative values in the $3$$\times$$3$ neighborhood of each local maximum of $s$.
%
%At the local maxima of $s$, we perform a sub-pixel corner position refinement by adding 
%\[dx=\frac{f_yf_{xy}-f_xf_{yy}}{f_{xx}f_{yy}-f_{xy}^2},\quad dy=\frac{f_xf_{xy}-f_yf_{xx}}{f_{xx}f_{yy}-f_{xy}^2} \]
%to each pixel position \cite{laureano2015} around a local maximum of $s$ and by then computing the grayscale centroids \cite{shortis1994} of these positions in the neighborhood of each local maximum.

Note, that the goal of this paper is not to derive a new corner detector and to set a new state of the art in precision of checkerboard corner detection. Indeed, one could alternatively use other sub-pixel accurate checkerboard corner detectors such as pattern based \cite{huang2018,laureano2015}, Fourier \cite{zhang2017} or Radon transform based \cite{duda2018}, pyramid based \cite{abeles2021} or neural network based approaches \cite{chen2018,chen2021,chen2023,kang2021a,kang2021b,wu2021}. Such filters can directly be applied to PuzzleBoards as well and may significantly improve corner detection acuracy. We use a Hessian-based local filter instead for two reasons: First, this is a good compromise between precision and algorithmic complexity, so the computation is reasonably fast. Second, the Hessian provides the basis for other relevant values such as the Eigenvector direction which is used in further processing steps. Moreover, this paper shows, that the position code can be reliably read even when using a simple corner detection algorithm.

\subsection{Neighbor Identification and Grid Reconstruction}

Once the checkerboard corner points have been detected, neighboring corners must be connected to reconstruct the grid. Typical approaches for structure recovery assume that all points of a rectangular grid \cite{geiger2012,itseez2015} or at least a sufficiently large rectangular subgrid is visible \cite{chen2021}. Others are more flexible by merging local Delaunay triangles \cite{laureano2015} or by traversing the connecting edges \cite{fuersattel2016}. All these methods have their limitations with strong occlusions or certain camera poses (e.g.\ even with orthogonal projection, checkerboard edges are not guaranteed to be part of the Delaunay triangulation if the projection angle is $66^\circ$ or larger). A more complex heuristic uses local grouping \cite{dao2010}, but cannot be applied to a PuzzleBoard at low resolutions since it requires locally straight edges.

We use a more robust two-step approach. First, for each corner point we detect the up to four direct neighbors in the grid by neighborhood search and second, we connect them  using a spanning tree based merging algorithm:

Neighbor Detection: Assuming a pinhole camera at a sufficient distance and a projection angle of at most $71.955^\circ$, all four direct grid neighbors of a checkerboard corner point $P$ are among its 9 nearest neighbors after projection. So we first determine the 9 nearest neighbors of each corner in parallel. Using the corner orientation given by the direction of the Hessian Eigenvectors, the direct neighbors (grid distance $(1,0)$) can be separated from the diagonal neighbors (grid distance $(1,1)$). %Furthermore, indirect neighbors $P_1,P_2$ with grid distance $(2,1)$ that have the same corner orientation as direct neighbors, are filtered out by checking whether \[\abs{g_{\frac16}-g_{\frac56}}\leq \abs{g_{\frac12}-\frac{g_0+g_1}2},\] where $g_0,g_{\frac16},g_{\frac12},g_{\frac56},g_1$ are the quadratic fit gray values at the sub-pixel positions \[P_1+(P_2-P_1)\cdot\left\{0, \frac16, \frac12, \frac56, 1\right\}.\] 
%However, neighbors at a grid distance of $(2,1)$ or $(3,2)$ cannot directly be filtered out using the Eigenvectors. 
The remaining nearest neighbor is chosen as the direct neighbor $+X$ and its counterpart $-X$ is the nearest in approximately the opposite direction. As the Hessian Eigenvectors at a corner point are the bisecting angles of the black and white sectors, the directions of the remaining two direct neighbors $Y+$ and $Y-$ are then derived by reflecting $X+$ over the Eigenvectors. This method is straight forward to compute and works for arbitrary projection angles as long as all direct neighbors are among the 9 nearest ones.

Grid Construction: After finding the up to four potential direct neighbors for each grid point, the corner points have to be connected to form a grid. For each edge, we compute a weight based on its length and the detection response of its endpoints, and use these weights to construct a minimal spanning forest. The forest structure automatically eliminates potential contradictions. We use Kruskal's algorithm with a union-find data structure, because it has only slightly above linear complexity after the initial sorting of the edges. Furthermore, the union-find structure guarantees that when two sub-trees are merged by a connecting edge, their relative orientation remains consistent. 

\subsection{Position Decoding}

Each code bit is being read using a local threshold by comparing the average grayscale value (at sub-pixel position) of two neighboring corner points with the sub-pixel grayscale value at the center point between them.

Since the patterns $\mathrm{A}$ and $\mathrm{B}$ are pseudorandom arrays \cite{etzion1988}, decoding can be done by measuring the cross-correlation of the observed pattern (after combining the bit repetitions to a single bit by majority voting) with the correct pattern. This allows to find the correct position and orientation of a visible part of a PuzzleBoard by computing 8 cross correlations of patterns of maximum size $3$$\times$$167$: each horizontal and vertical bit pattern with each of the two patterns $\mathrm{A}$ and $\raisebox{\depth}{\rotatebox{90}{B}}$ and their $180^\circ$ rotated versions $\raisebox{\depth}{\rotatebox{180}{A}}$ and $\raisebox{\depth}{\rotatebox{270}{B}}$. Then, the exact position can be computed by using modulo arithmetic. For example, if the highest cross correlation of the detected horizontal and vertical bit pattern occurs with the pattern pair $\mathrm{A}$ and $\raisebox{\depth}{\rotatebox{90}{B}}$ at positions \[x_A,y_A,\quad x_B,y_B\] with \[x_A,y_B\in\{0\ldots 166\} \text{ and } x_B,y_A\in\{0\ldots 2\},\] then the detected code begins at position 
\[\left(\begin{array}{c}x_A+167\cdot[x_A-x_B\pmod{3}]\\ y_B+167\cdot[y_B-y_A\pmod{3}]\end{array}\right)\begin{array}{c}\\.\end{array}\]

A measure of error correction ability is the Hamming distance of a pattern to the code for any incorrect position or rotation: As the orientation of the perceived pattern in the complete code is not known, all 4 possible orientations need to be checked. A sufficiently large pattern would result in two codes of size $3$$\times$$167$ and $167$$\times$$3$, which need to be cross correlated with the base codes $A$ and $B$ and their rotations. If such codes have been read without errors, they are cyclic translations of one of the base patterns or their rotated versions and can thus be associated with $\mathrm{A}$, $\raisebox{\depth}{\rotatebox{180}{A}}$, $\mathrm{B}$ and $\raisebox{\depth}{\rotatebox{180}{B}}$. At correct rotation and translation, the cross correlation as well as the number of equal bits is thus $3\cdot167=501$. For each other rotation and/or translation, the following tables shows the maximal number of equal bits for patterns $\mathrm{A}$ and $\mathrm{B}$:
\[
\begin{array}{c|c|c|c|c|}
& \mathrm{A} & \raisebox{\depth}{\rotatebox{180}{A}} & \raisebox{\depth}{\rotatebox{90}{B}} & \raisebox{\depth}{\rotatebox{270}{B}} \\
\hline
\mathrm{A} & 53 & 93 & 21 & 19 \\
\hline
\end{array}
\qquad
\begin{array}{c|c|c|c|c|}
& \mathrm{B} & \raisebox{\depth}{\rotatebox{180}{B}} & \raisebox{\depth}{\rotatebox{90}{A}} & \raisebox{\depth}{\rotatebox{270}{A}} \\
\hline
\mathrm{B} & 58 & 105 & 19 & 21 \\
\hline
\end{array}
\]

This means, that the minimal Hamming distance for false rotations and/or translations is given by:

\[
\begin{array}{c|c|c|c|c|}
& \mathrm{A} & \raisebox{\depth}{\rotatebox{180}{A}} & \raisebox{\depth}{\rotatebox{90}{B}} & \raisebox{\depth}{\rotatebox{270}{B}} \\
\hline
\mathrm{A} & 448 & 408 & 480 & 482 \\
\hline
\end{array}
\qquad
\begin{array}{c|c|c|c|c|}
& \mathrm{B} & \raisebox{\depth}{\rotatebox{180}{B}} & \raisebox{\depth}{\rotatebox{90}{A}} & \raisebox{\depth}{\rotatebox{270}{A}} \\
\hline
\mathrm{B} & 443 & 396 & 482 & 480 \\
\hline
\end{array}
\]

As the patterns always occur in certain pairs, we get the following minimal Hamming distances for the possible combinations of patterns:
\[
\begin{array}{c|c|c|c|c|}
& \mathrm{A},\mathrm{B} & 
\raisebox{\depth}{\rotatebox{180}{A}}, \raisebox{\depth}{\rotatebox{180}{B}} &
\raisebox{\depth}{\rotatebox{90}{B}}, \raisebox{\depth}{\rotatebox{90}{A}} &
\raisebox{\depth}{\rotatebox{270}{B}}, \raisebox{\depth}{\rotatebox{270}{A}} \\
\hline
\mathrm{A},\mathrm{B} & 891 & 804 & 962 & 962 \\
\hline
\end{array}
\]

The minimal Hamming distance at wrong position is thus $804$ bit. Thus, the second highest cross-correlation occurs at a position where only 198 out of 1002 bits are the same. Since $804/2=402$, for a sufficiently large calibration board up to $401/1002\approx 40\%$ of all bits are allowed to be decoded incorrectly \emph{after} averaging over all repetitions and thus the error rate per edge may be even higher.

\begin{figure}
\centering
\includegraphics[width = 0.32\textwidth]{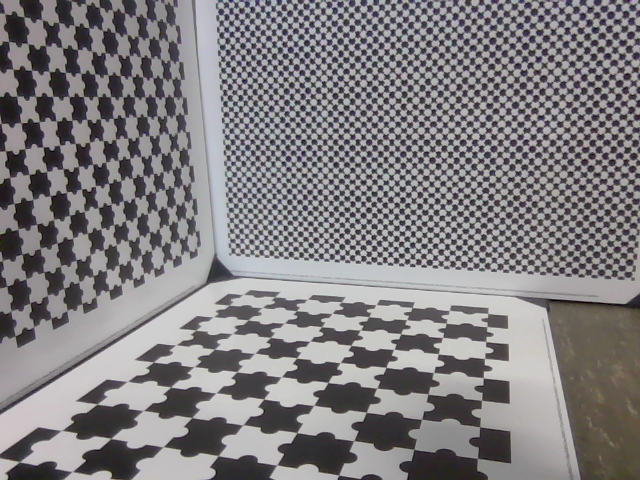}
\includegraphics[width = 0.32\textwidth]{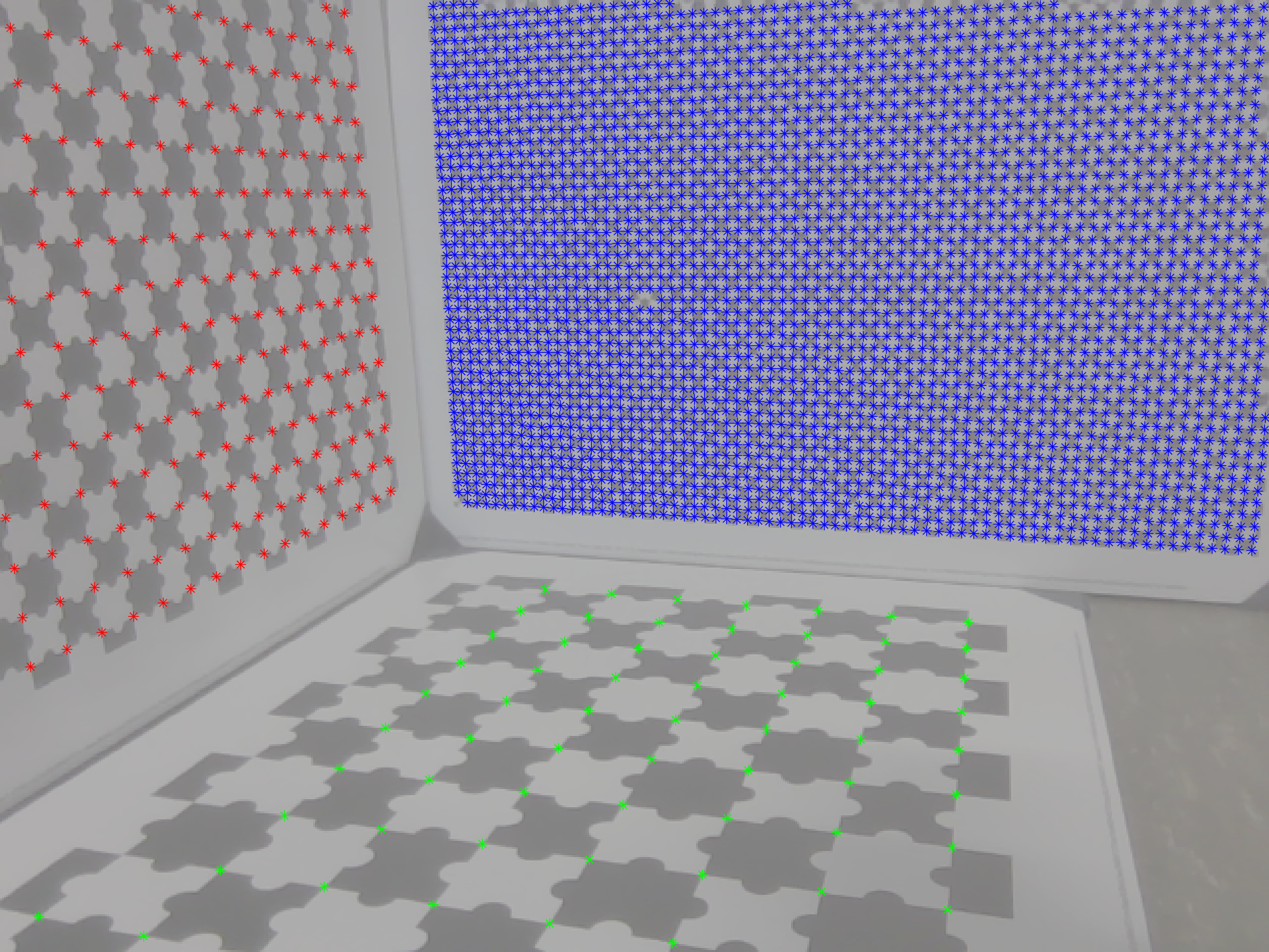}
\includegraphics[width = 0.236\textwidth]{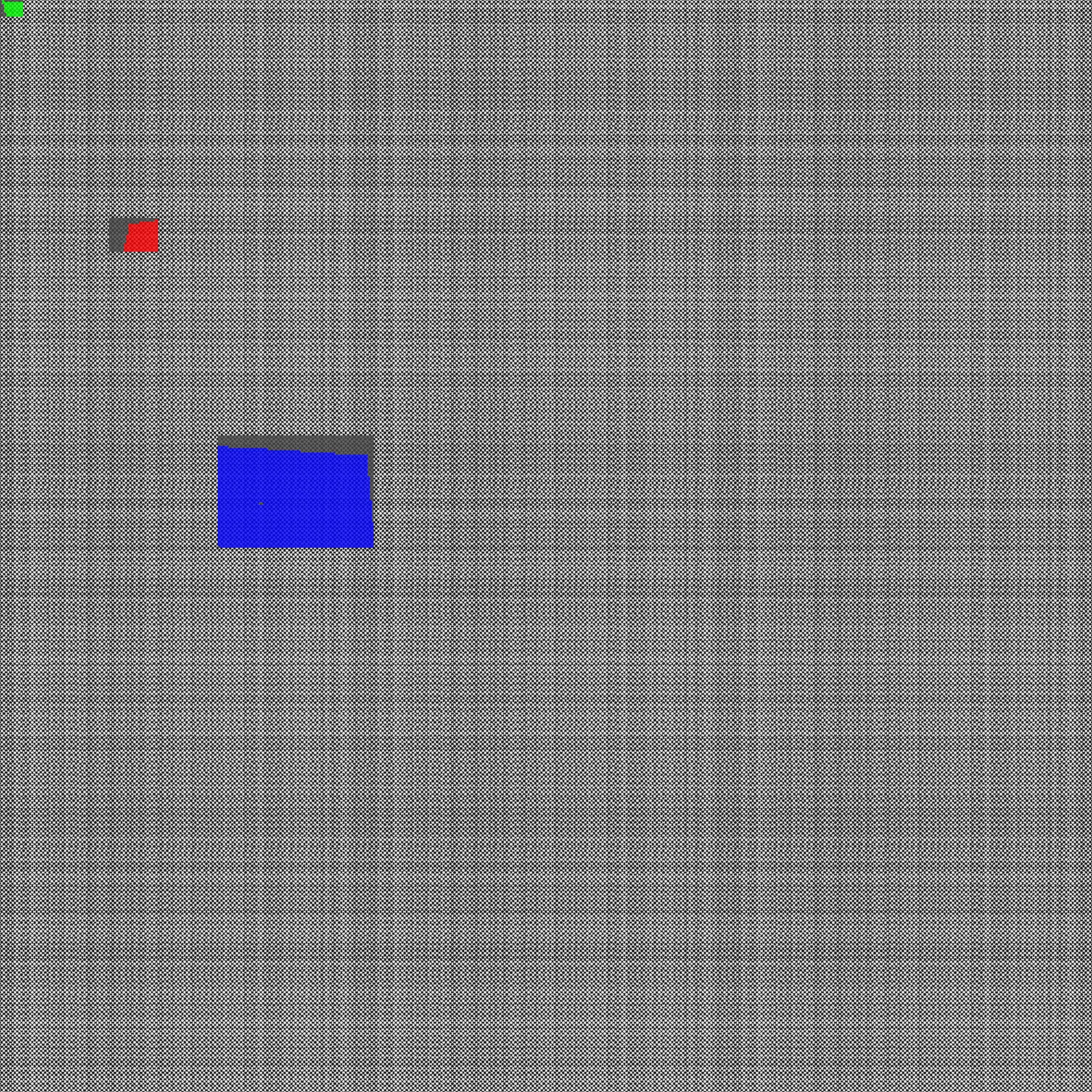}\\
\caption{Left: Sample image showing parts of three different PuzzleBoard targets. Center: Detected and successfully decoded corner points (red, green, blue). Right: Detected corners (red, green, blue) and their position in the total pattern of size $501$$\times$$501$ (use zoom in digital paper). The invisible or not detected corner points of the three calibration boards are shown in gray.}
\label{fig:threetargets}
\end{figure}

\section{Experiments} \label{sec3}

We printed three calibration targets of size $7$$\times$$10$, $15$$\times$$22$ and $51$$\times$$71$ pieces. As these show different parts of the total pattern of $501$$\times$$501$ pieces and as the decoding algorithm is able to detect more than one connected pattern at a time, we are able to decode several targets simultaneously in one image, as shown in Figure~\ref{fig:threetargets}.

\begin{figure}
\centering
\includegraphics[width = 0.32\textwidth]{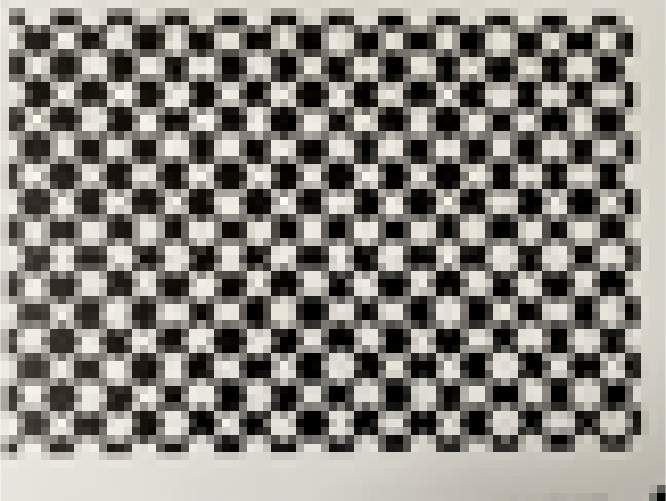}
\includegraphics[width = 0.32\textwidth]{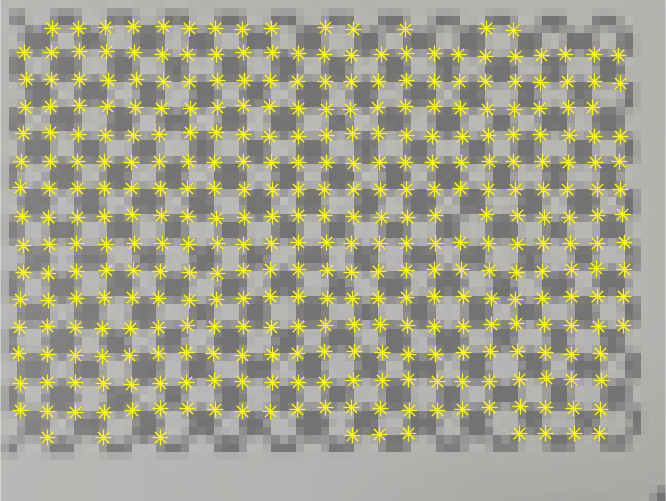}
\includegraphics[width = 0.32\textwidth]{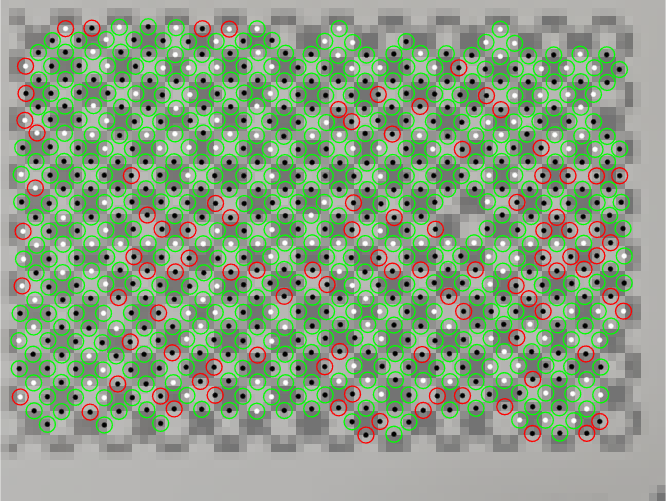}\\
\includegraphics[width = 0.32\textwidth]{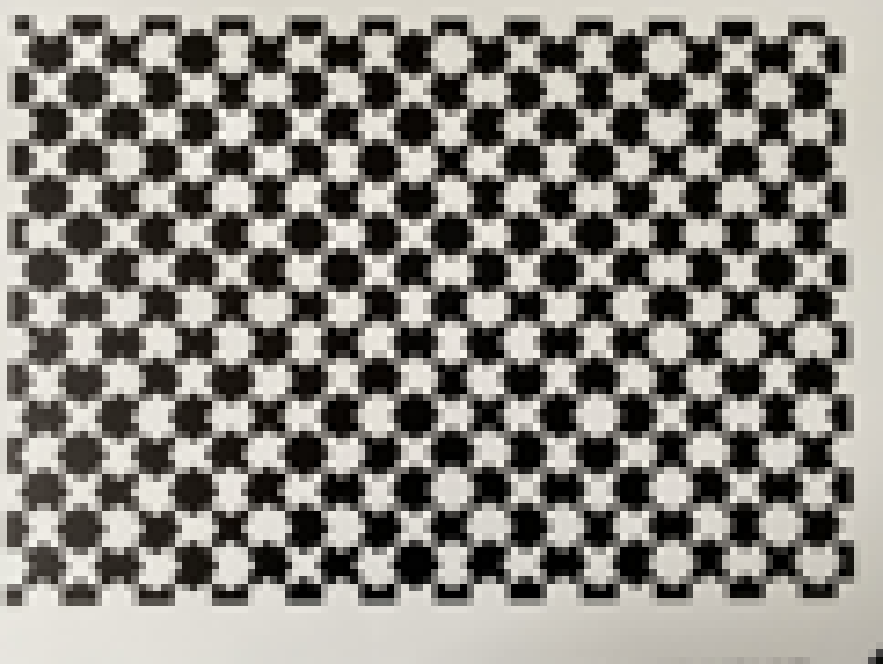}
\includegraphics[width = 0.32\textwidth]{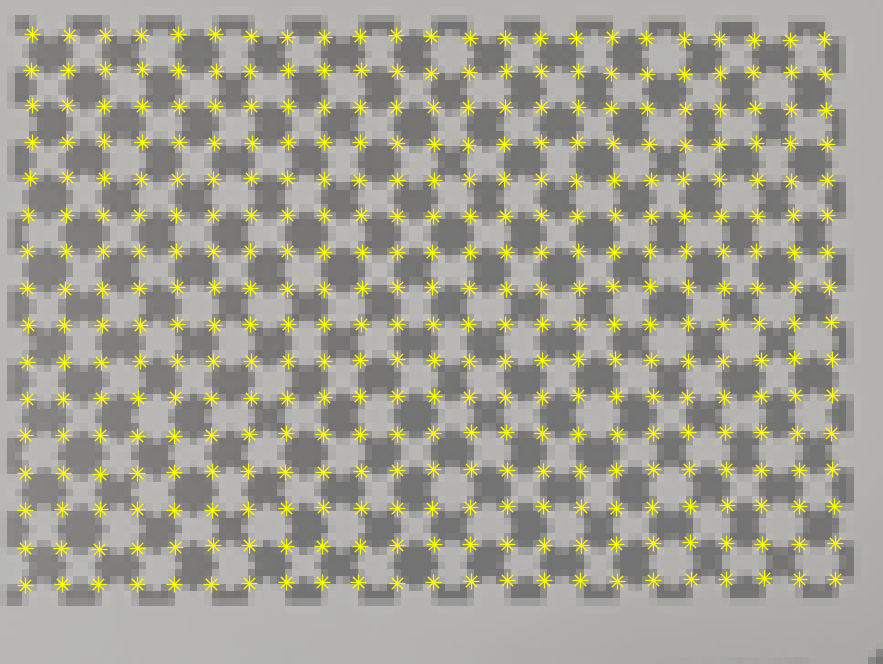}
\includegraphics[width = 0.32\textwidth]{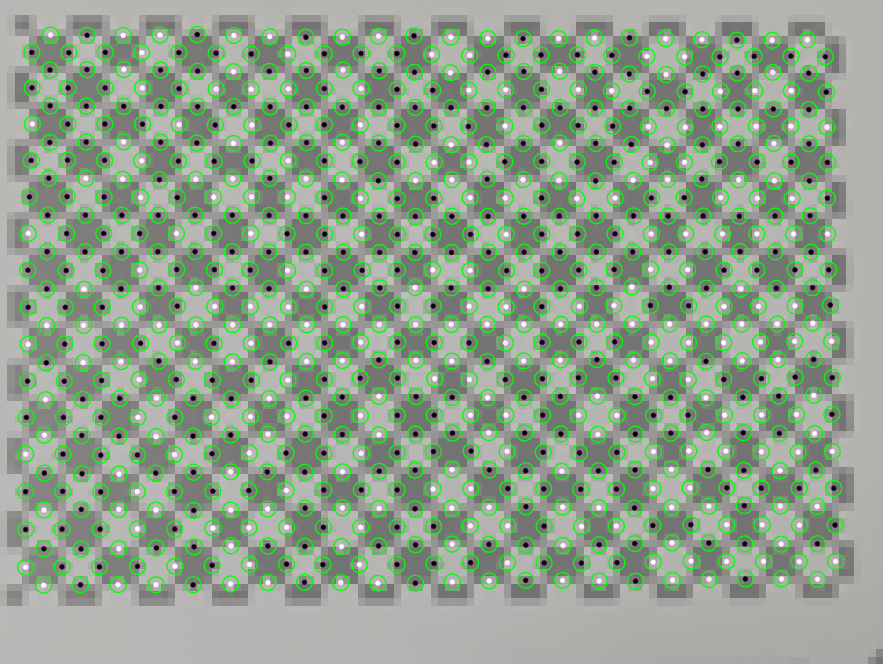}\\
\includegraphics[width = 0.32\textwidth]{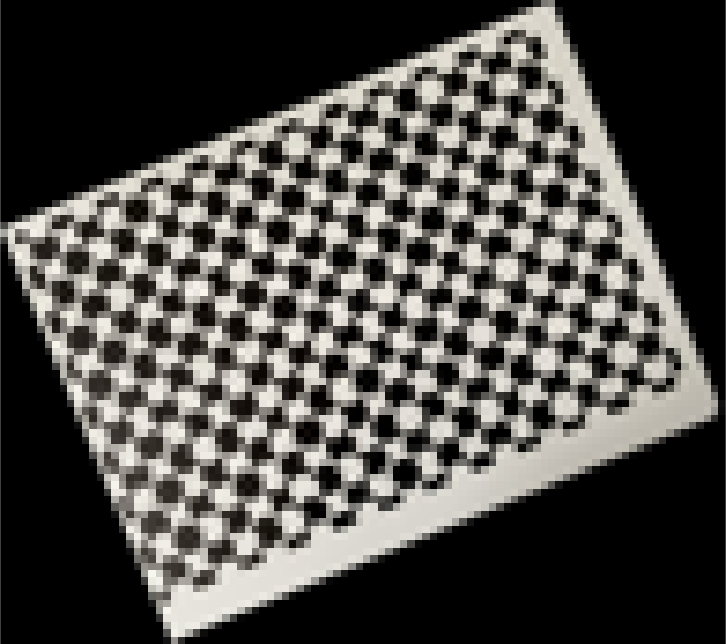}
\includegraphics[width = 0.32\textwidth]{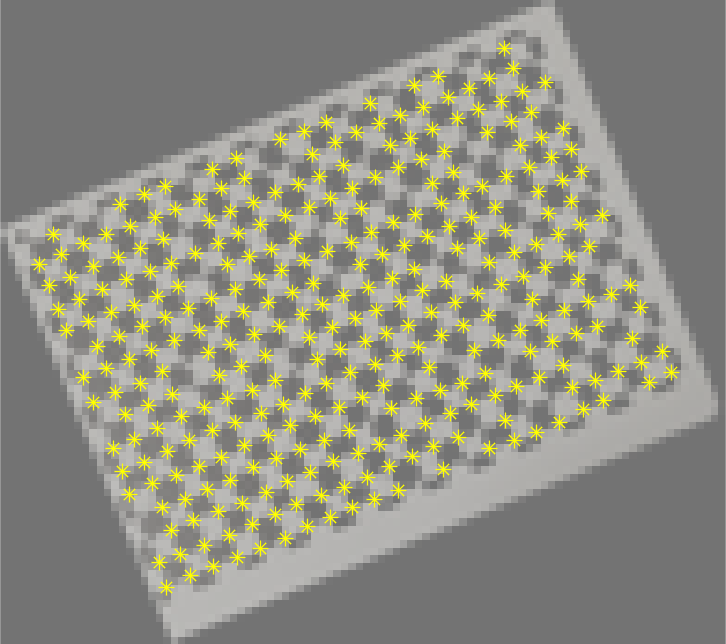}
\includegraphics[width = 0.32\textwidth]{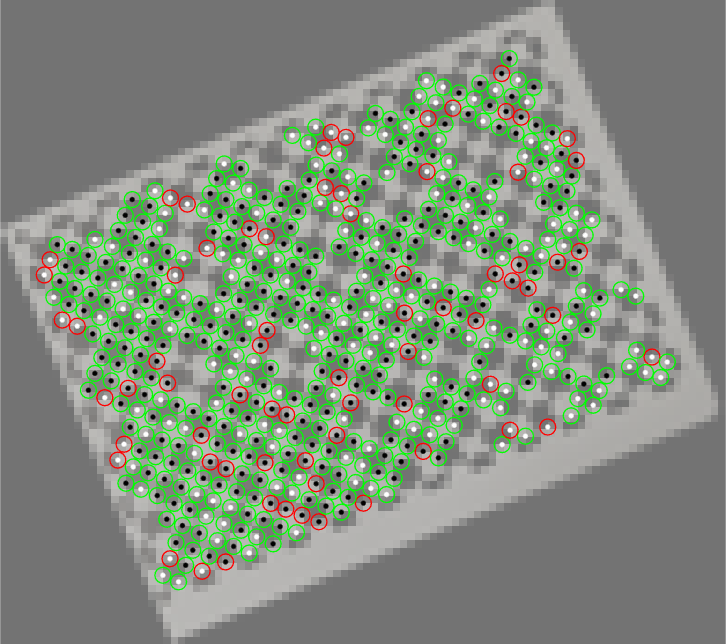}\\
\includegraphics[width = 0.32\textwidth]{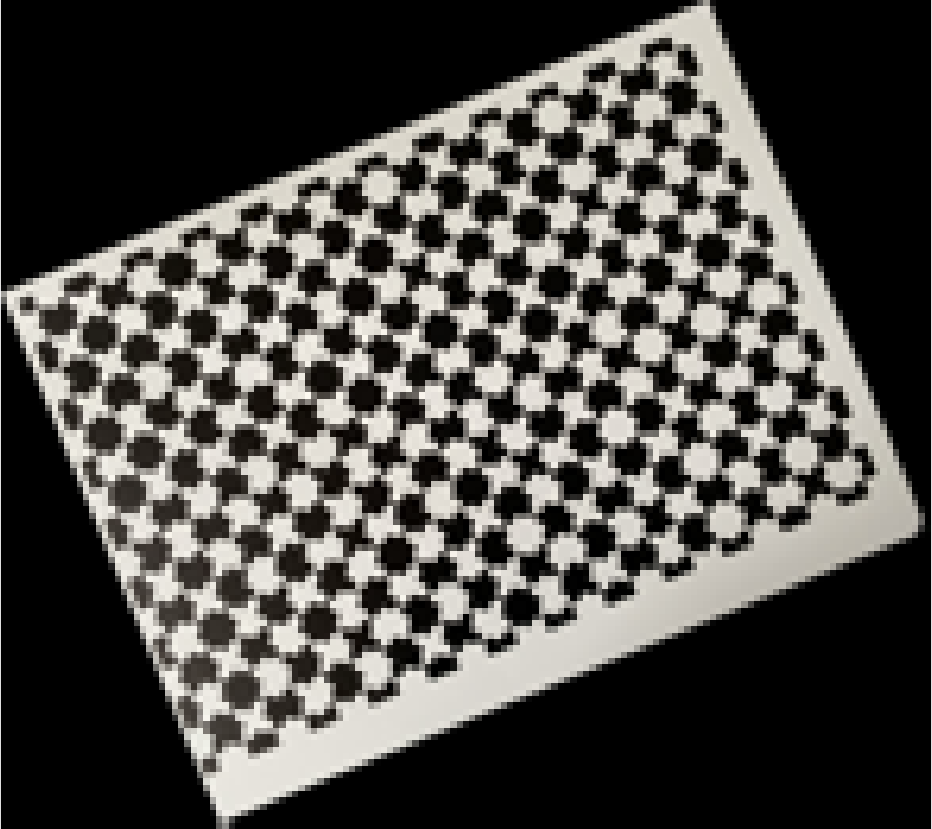}
\includegraphics[width = 0.32\textwidth]{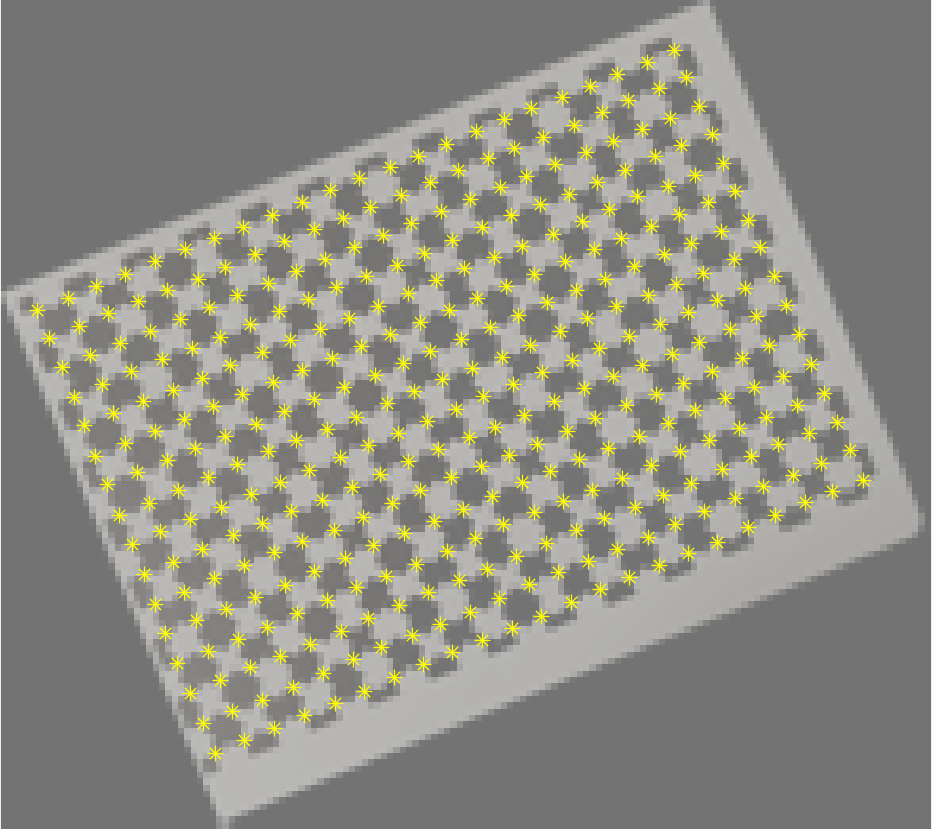}
\includegraphics[width = 0.32\textwidth]{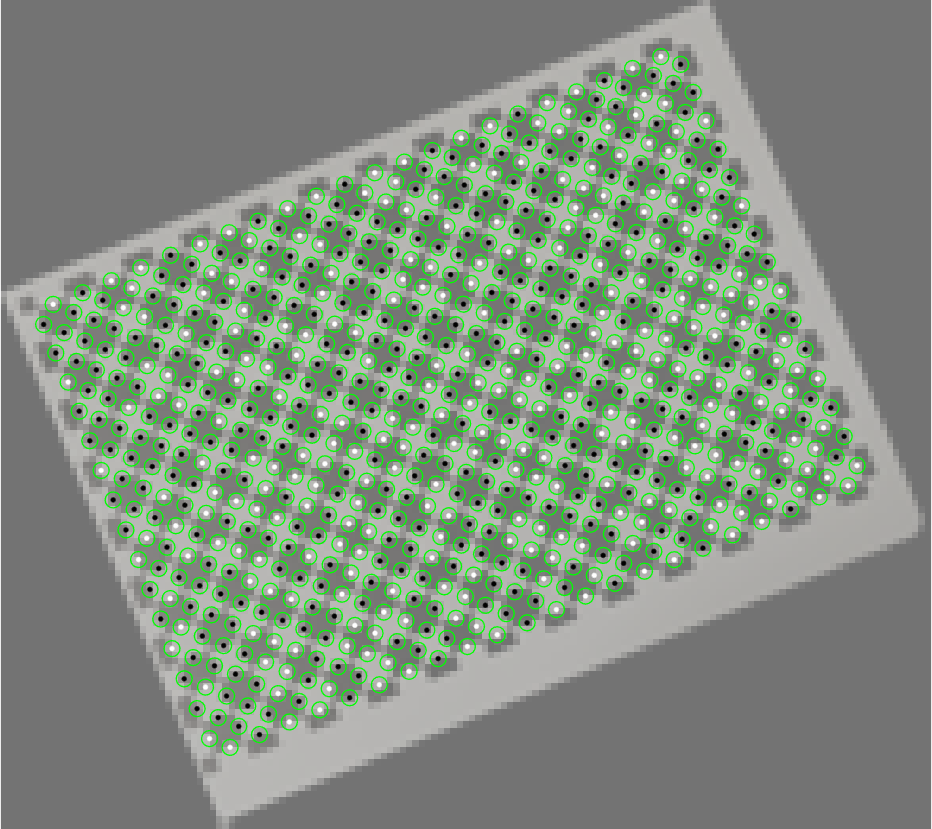}\\
\caption{Left: Downsampled images of a PuzzleBoard of size $22$$\times$$15$ pieces. Center: Detected sub-pixel corner points. Right: Edge decodings (green: correct; red: incorrect). The incorrect decodings are automatically identified by the error correction. First row: resolution 3.33 pixels per edge. 341 out of 368 corners are detected and 633 out of 697 edges are decoded with 537 of them being decoded correctly. Second row: resolution 5 pixels per edge. All corners are detected and all edges are decoded correctly. Third row: resolution 3.33 pixels per edge at rotation $22.5^\circ$. 312 out of 368 corners are detected and 514 out of 697 edges are decoded with 443 of them being decoded correctly. Fourth row: resolution 5 pixels per edge at rotation $22.5^\circ$. All corners are detected and all edges are decoded correctly. In all cases, the correct code position of each detected corner point could be derived due to the error correction.}
\label{fig:lowres}
\end{figure}

We tested the robustness regarding low-resolution images by downsampling a high-resolution image of a PuzzleBoard calibration target before decoding (see Figure~\ref{fig:lowres}). Even with a resolution of 5 pixels per PuzzleBoard edge, the entire grid could be decoded without any decoding errors. With an edge length of 3.33 pixels most of the grid corners could still be detected and all their code positions could be correctly derived, although there are approximately $15\%$ code bit errors. In comparison, reliably decoding the $7$$\times$$7$ ArUco markers of a ChArUco board typically requires a checkerboard edge length of at least 25 pixels.

We implemented the decoding algorithm in C++. For performance evaluation, we ran the algorithm on  one core of a 2.80GHz Intel Core i7-8569U CPU. We measured the processing time including corner point detection, grid construction and position decoding and compare it to the OpenCV detector findChessboardCornerSB \cite{duda2018,itseez2015} which is known for being fast on large images \cite{hillen2023}.

\begin{figure}
\centering
\includegraphics[width = 0.49\textwidth]{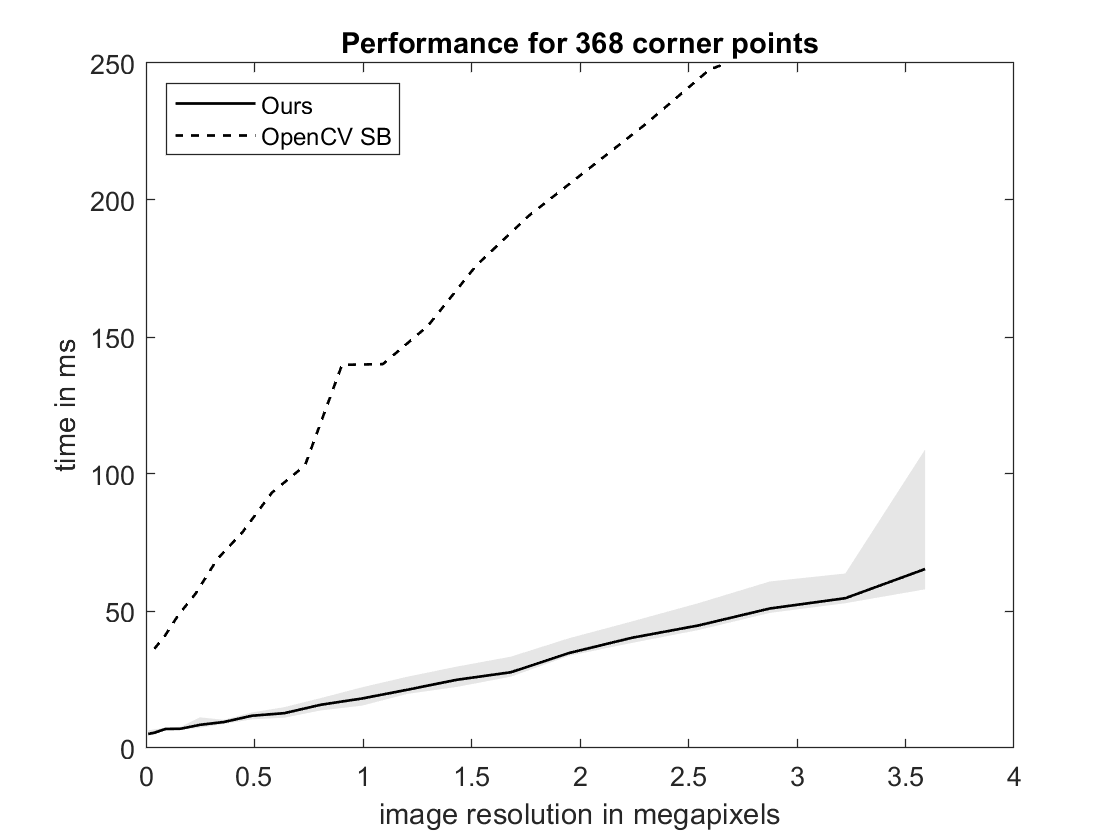}
\includegraphics[width = 0.49\textwidth]{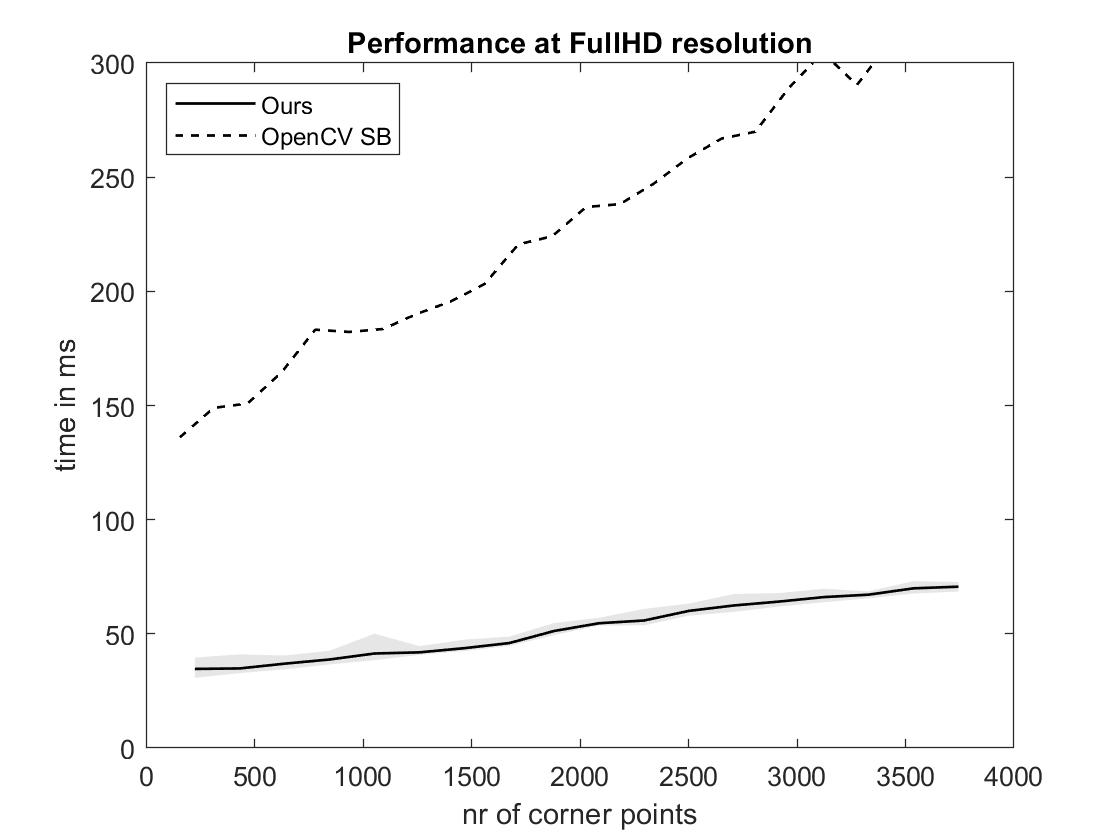}
\caption{Runtime dependency on image resolution (left) and pattern size (right)}
\label{fig:perf}
\end{figure}

For the first performance experiment, we took images of the $15$$\times$$22$ PuzzleBoard calibration target (which contains $16$$\times$$23=368$ corner points) at different resolutions ranging from $116$$\times$$87$ to $2188$$\times$$1640$ pixels. Figure~\ref{fig:perf}(left) shows the average processing time  for 10 measurements per resolution, as well as the interval from minimum to maximum processing time. As can be seen, the algorithm scales linearly with image size, and still runs at 15fps for 3.5 megapixel images.

For the second experiment, we took images of a $51$$\times$$71$ PuzzleBoard calibration target with $52$$\times$$72=3744$ corner points at FullHD resolution and varied the number of visible corner points by covering parts of the target. As Figure~\ref{fig:perf}(right) shows, the algorithm scales approximately linearly in the number of grid points and runs with 29fps for 225 corner points and 14fps for the whole 3744 corner points.
As can be seen, our algorithm is significantly faster than the OpenCV detector findChessboardCornerSB \cite{duda2018,itseez2015}.

%\subsection{Puzzle Markers}
%\subsection{Cylindrical Targets}

\section{Conclusion and Future Work} \label{sec4}

We introduced the PuzzleBoard, a new calibration pattern which combines the precision of checkerboard patterns with a lightweight position encoding scheme, which allows to derive the identity of each checkerboard corner point from a local neighborhood of just $3$$\times$$3$ puzzle pieces. As the necessary neighborhood is such small, the method works even under severe occlusion. The code can be read at extremely low resolutions, which makes the pattern suitable not only for calibration purposes but also for marker based camera localization with e.g.\ low resolution cameras on embedded devices or at large distances.
We further presented a fast algorithm for detecting and decoding PuzzleBoard patterns. In future, we plan to optimize the algorithm for a wider range of projection angles and to investigate the robustness when using PuzzleBoard patches as fiducial markers.

\end{document}